\documentclass[11pt]{article}

\usepackage[preprint]{acl}
 
\usepackage{times}
\usepackage{latexsym}

\usepackage[T1]{fontenc}

\usepackage[utf8]{inputenc}
\usepackage{amsmath}

\usepackage{microtype}

\usepackage{inconsolata}

\usepackage{graphicx}
\graphicspath{{figures/}{figures/benchmark/}}

\usepackage[table]{xcolor}
\usepackage{booktabs}
\usepackage{colortbl}
\usepackage{multirow}
\usepackage{array}
\usepackage{tabularx}
\usepackage{makecell}
\usepackage{arydshln}
\usepackage{amssymb}
\usepackage{pifont}
\usepackage{subcaption}
\usepackage{enumitem}
\usepackage{placeins}
\usepackage{listings}
\usepackage[most]{tcolorbox}

\lstdefinelanguage{json}{
  morestring=[b]",
  morekeywords={true,false,null},
  sensitive=false,
  showstringspaces=false
}

\definecolor{promptbg}{RGB}{237,243,250}
\definecolor{promptframe}{RGB}{200,215,235}
\definecolor{catgreen}{RGB}{230,245,230} 
\definecolor{catblue}{RGB}{230,240,255}  
\definecolor{catyellow}{RGB}{255,250,225}
\definecolor{catred}{RGB}{255,230,230}   

\newcommand{\promptbox}[1]{%
  \begingroup
  \setlength{\fboxsep}{6pt}%
  \setlength{\fboxrule}{0.5pt}%
  \fcolorbox{promptframe}{promptbg}{%
    \begin{minipage}{0.97\linewidth}
      \scriptsize\ttfamily\raggedright
      \setlength{\parskip}{2pt}%
      #1
    \end{minipage}%
  }%
  \endgroup
}

\title{{WindowsWorld}: A Process-Centric Benchmark of Autonomous GUI Agents in Professional Cross-Application Environments}

\author{
Jinchao Li\textsuperscript{1,2},
Yunxin Li\textsuperscript{1,2},
Chenrui Zhao\textsuperscript{1},
Zhenran Xu\textsuperscript{1},
Baotian Hu\textsuperscript{1,2}\thanks{Corresponding author.},
Min Zhang\textsuperscript{1,2} \\
\textsuperscript{1}Harbin Institute of Technology, Shenzhen \\
\textsuperscript{2}Shenzhen Loop Area Institute \\
\texttt{jinchaoli@slai.edu.cn \quad xuzhenran@stu.hit.edu.cn} \\ 
\texttt{qq1836143240@outlook.com \quad \{liyx, hubaotian, 
 zhangmin2021\}@hit.edu.cn
 }}

\begin{document}
\maketitle
\begin{abstract}
While GUI agents have shown impressive capabilities in common computer-use tasks such as OSWorld, current benchmarks mainly focus on isolated and single-application tasks. This overlooks a critical real-world requirement of coordinating across multiple applications to accomplish complex profession-specific workflows. To bridge this gap, we present a computer-use benchmark in cross-application workflows, named \textbf{WindowsWorld}, designed to systematically assess GUI Agents on complex multi-step tasks that mirror real-world professional activities. Our methodology uses a multi-agent framework steered by 16 occupations to generate four difficulty-level tasks with intermediate inspection, which are then refined by human review and executed in a simulated environment. The resulting benchmark contains 181 tasks with an average of 5.0 sub-goals across 17 common desktop applications, of which 78\% are inherently multi-application. Experimental results of leading large models and agents show that: 1) All computer-use agents perform poorly on multi-application tasks ($<$ \textbf{21\% success rate}), far below the performance of simple single-app tasks;
2) They largely fail at tasks requiring conditional judgment and reasoning across $\geq$3 applications, stalling at early sub-goals;
3) Low execution efficiency, where tasks often fail despite far exceeding human step limits. Code, benchmark data, and evaluation resources are available at \href{https://github.com/HITsz-TMG/WindowsWorld}{\texttt{github.com/HITsz-TMG/WindowsWorld}}.

\end{abstract}

\section{Introduction}


\begin{figure}[t]
    \centering
    \begin{subfigure}[b]{0.48\columnwidth}
        \centering
        \includegraphics[width=\linewidth]{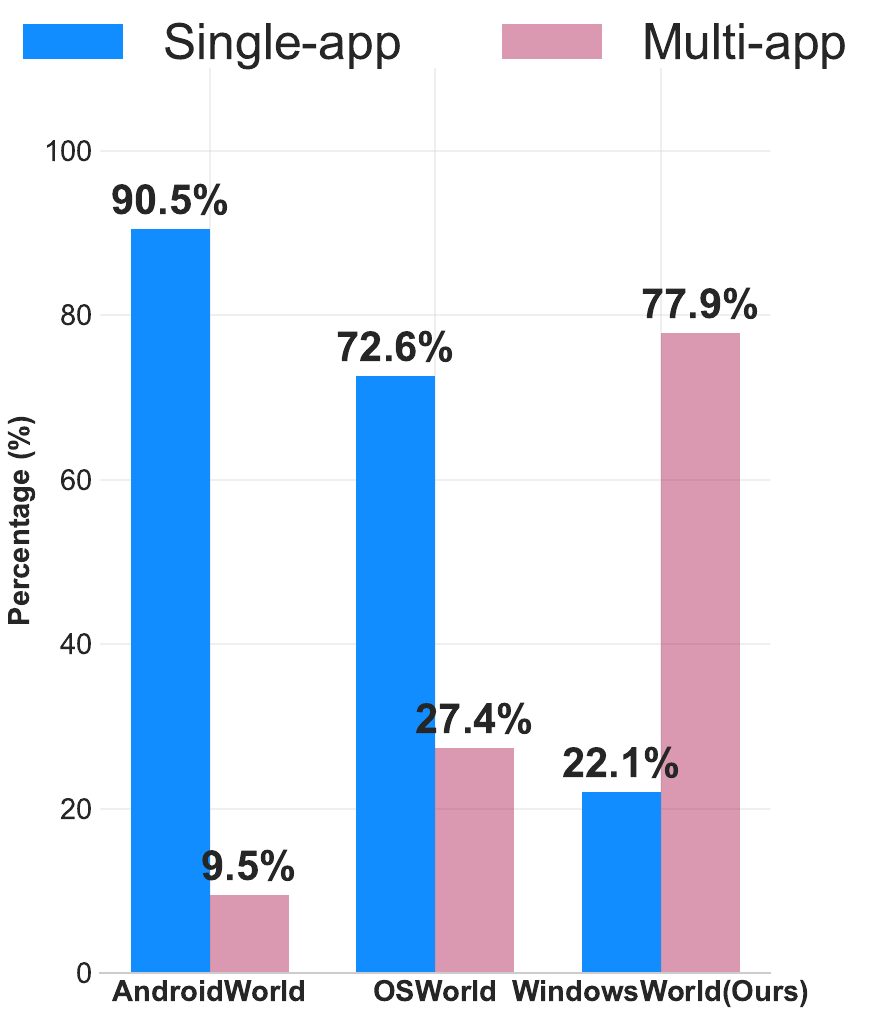}
        \caption{}
        \label{fig:task-distribution}
    \end{subfigure}
    \hfill
    \begin{subfigure}[b]{0.49\columnwidth}
        \centering
        \includegraphics[width=\linewidth]{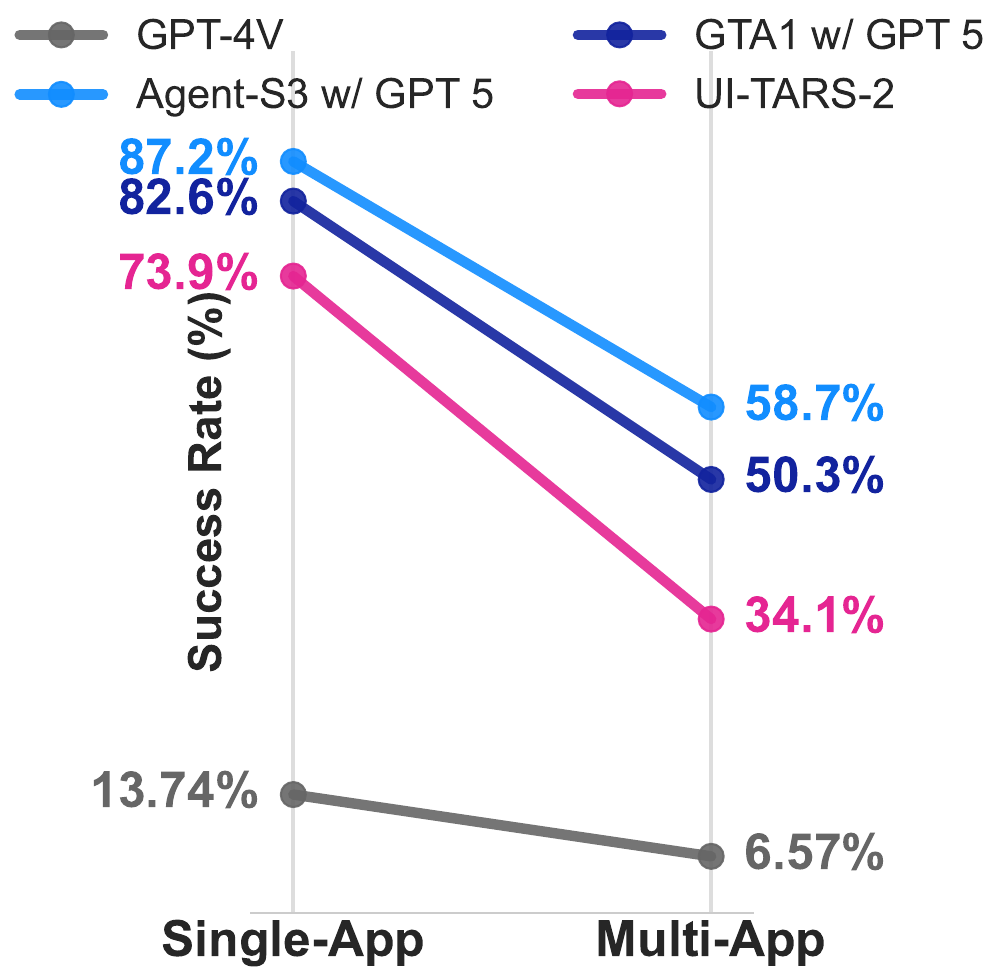}
        \caption{}
        \label{fig:success-drop}
    \end{subfigure}
    \caption{
    \textbf{Limitations of current GUI benchmarks in real-world tasks.}
    (a) Distribution of single/cross-application tasks across benchmarks.
    (b) The task success rate drops with the increase in applications.
    }
    \label{fig:benchmark-limitations}
    \vspace{-0.5em}
\end{figure}

Autonomous Computer-Use Agents (CUAs) \citep{assouel_unsolved_2023,chen_gui-world_2024,qin2025ui,sager2025comprehensive}, which interpret natural language instructions to interact with digital environments by directly controlling user interfaces, represent a pivotal step toward Artificial General Intelligence (AGI). Recent progress in large multimodal reasoning models suggests that GUI agents increasingly rely on coupled perception, reasoning, and planning abilities rather than pure visual grounding alone \citep{li2025perception}. Early research \citep{li2025screenspot, rawles2023androidinthewild, deng2023mind2web, kapoor2024omniact} was primarily confined to static offline environments, focusing on single-step action prediction or limited task completion within isolated application screens. A significant shift has occurred with the advent of execution-based evaluation in online operating systems, enabled by scalable Virtual-Machine (VM) environments such as OSWorld \citep{xie2024osworld} and Windows Agent Arena \citep{bonatti2024windows}.
However, these benchmarks are still failing to capture the cognitive demands of real-world professional work that requires using multiple applications. This gap is becoming increasingly consequential as language-centric omnimodal foundation models continue to expand their modality coverage and general interaction capabilities \citep{li2025uni}.

Specifically, existing benchmarks suffer from three critical limitations. \textit{Firstly, cross-application tasks remain insufficient}, where most benchmarks focus on single-app navigation or short-horizon workflows, e.g., OSWorld \citep{xie2024osworld} and AndroidWorld \citep{rawles2024androidworld}, under-representing the cross-application coordination required in professional settings. As shown in Figure~\ref{fig:benchmark-limitations}(a), widely used benchmarks exhibit a strong imbalance in task composition, with single-application tasks accounting for the majority of instances, while multi-application tasks remain sparsely represented ($<$28\%). And GUI agents demonstrate high success rates on single-application tasks, yet their performance deteriorates markedly as task complexity increases. Figure~\ref{fig:benchmark-limitations}(b) shows a steep drop in success rate when tasks require coordination across multiple applications, highlighting a significant gap between benchmark difficulty and real-world demands. 
\textit{Secondly, the ``all-or-nothing'' scoring paradigm}, where judging success solely by the final success without progress evaluation leads to a lack of diagnostic granularity \citep{kong2025mobileworld, yang2025probench}. In high-difficulty tasks where most state-of-the-art models fail to reach the final goal, current metrics (SR) cannot distinguish between models that made significant partial progress and those that failed at the first step of tasks. \textit{Thirdly, benchmark construction remains bottlenecked by labour-intensive manual curation or repetitive template substitution}. Traditional pipelines  \citep{xie2024osworld} typically depend on forum crawling followed by extensive human editing, which is unscalable and frequently lacks the human-centric context of authentic professional routines \citep{mu2025gui}. Furthermore, these benchmarks fail to automatically synthesize the complex file dependencies required to initialize evaluation tasks, creating a gap between simulated environments and real-world productivity.

To bridge these critical gaps, we introduce \textbf{WindowsWorld}, a comprehensive, challenging benchmark built on high-fidelity Windows virtual machines. It is constructed by a human-centric multi-agent data automation generation followed by human review. Specifically, it consists of:
\begin{itemize}[leftmargin=*]
    \item \textbf{Professional-Grade Multi-APP Tasks}: 
    WindowsWorld comprises 181 tasks spanning 17 desktop applications, systematically generated based on 16 distinct personas across 5 categories. These tasks are organized into four difficulty levels that progressively increase planning horizon and cross-application coordination, including infeasible tasks to evaluate goal recognition and abstention. The benchmark is dominated by about 78\% realistic multi-application workflows, capturing structured information transfer and common office productivity scenarios.
    \item \textbf{Fine-Grained Intermediate Process Checking}: Departing from only adopting final success evaluation metrics, we implement intermediate checking points for each task instruction. By validating essential sub-goals for each task, we assign partial-progress scores, providing higher discriminative power for evaluating computer-use agents' performance on long-horizon and high-difficulty workflows.
    \item \textbf{Automatic Task Construction}: We propose a human-in-the-loop multi-agent framework designed to generate real-world tasks. Initially, a Personas‑based Generator Agent creates scenarios and task instructions grounded in specific professional roles and daily workflows. This is followed by a Refiner Agent phase to eliminate redundant tasks, and an Environment Generator Agent phase that synthesizes the necessary files for each mission. Notably, before the final environment generation step, human evaluators select the required task instructions.
    \vspace{-2pt}
\end{itemize}

Extensive evaluations of leading GUI models and agents, e.g., Gemini-3-Pro \citep{team2025gemma}, GPT-5.2 \citep{team2025openai}, Agent S3 \citep{Agent-S3}, and others, reveal a significant performance gap in handling non-linear workflows and process-dependent constraints compared to simple tasks. The main contributions of our work are:
1) We present a comprehensive computer-use benchmark to assess GUI Agents' capability of completing multi-application collaborative tasks in a professional setting, with a progress evaluation metric besides the final success rate.
2) We introduce a human-centric multi-agent automatic generation pipeline, which significantly reduces the cost of dataset construction.
3) Experimental results reveal that current GUI agents perform well within a single application, yet their success rate and efficiency drop significantly when meeting tasks across applications. The best model Gemini-3-flash-preview achieves about only a 20\% success rate.


\section{Related Work}

\begin{figure*}[t]
  \centering
  \begin{minipage}[c]{0.38\textwidth}
    \centering
    \small
    \renewcommand{\arraystretch}{1.0} 
    \begin{tabular}{@{} l l r @{}}
      \toprule
      \textbf{Category} & \textbf{Applications} & \textbf{Count} \\
      \midrule
      
      \multirow{4}{*}{Office Software} & Word & 40 \\
                                       & Excel & 73 \\
                                       & PowerPoint & 18 \\ 
                                       & Acrobat & 11 \\ 
      \midrule 
                         
      Communication & Thunderbird & 72 \\ 
      \midrule
      
      Web Browser   & Chrome & 70 \\ 
      \midrule
      
      \multirow{4}{*}{System Manager}  & File Explorer & 59\\
                                       & Calculator & 7 \\ 
                                       & Task Manager & 2\\ 
                                       & Snipping Tool & 1 \\
      \midrule
                         
      \multirow{4}{*}{Multimedia}      & GIMP & 14 \\
                                       & Paint & 11 \\
                                       & Photos & 7 \\ 
                                       & VLC & 1 \\
      \midrule
                         
      \multirow{3}{*}{Programming}     & VS Code & 30 \\
                                       & PowerShell & 18 \\
                                       & Windows Terminal & 9 \\
      \bottomrule
    \end{tabular}
  \end{minipage}
  \hfill
  \begin{minipage}[c]{0.50\textwidth}
    \centering
    \includegraphics[width=\linewidth]{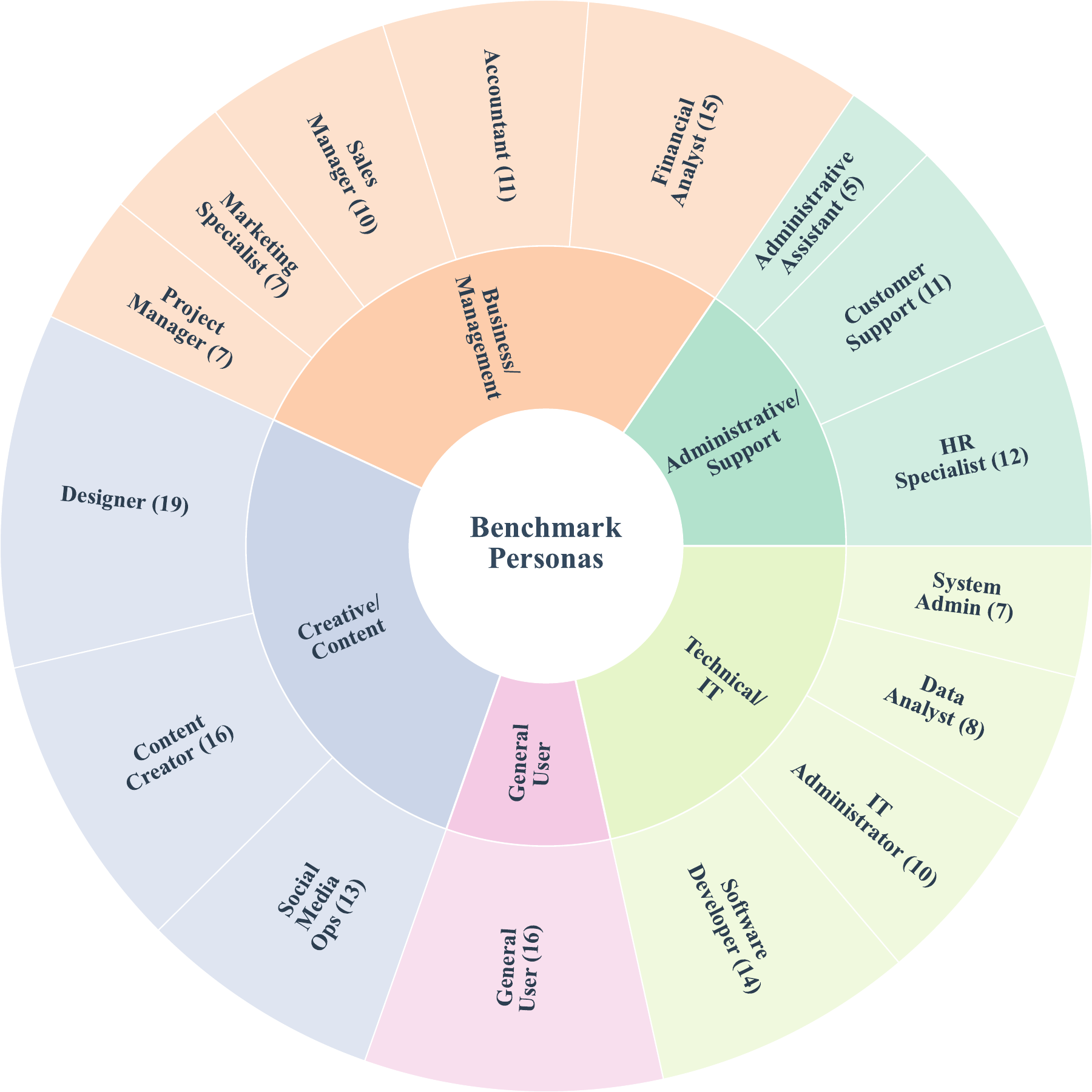}
  \end{minipage}

  \caption{\textbf{(Left)} Distribution of applications across different categories and their respective task counts. \textbf{(Right)} 16 Personas in WindowsWorld are categorized in \textbf{5} major domains, each with fine-grained personal roles.}
  \label{fig:table_left_image_right}
\end{figure*}

\textbf{Desktop OS Benchmarks}.
Desktop environments present unique challenges due to high-resolution displays and heterogeneous widget styles. Early benchmarks, e.g., MiniWoB \citep{shi2017world}, MiniWoB++ \citep{liu2018reinforcement}, WebShop \citep{yao2022webshop}, Mind2Web \citep{deng2023mind2web}, WebArena \citep{zhou2023webarena}, and VisualWebArena \citep{koh2024visualwebarena}, focus on the single-step accuracy of computer-use agents or simple tasks. OSWorld \citep{xie2024osworld} introduced the first scalable environment for Ubuntu, Windows, and macOS, supporting execution-based evaluation. Windows Agent Arena \citep{bonatti2024windows} expanded this with a focus on Windows OS and cloud-based parallelization to accelerate large-scale testing.  OSUniverse \citep{davydova2025osuniverse} introduced tasks with increasing complexity levels (from Paper to Gold) to benchmark the dexterity and precision of advanced agents. Our work builds upon the OSWorld infra but pushes the boundary of task complexity and diagnostic evaluation.

\textbf{Evaluation of GUI Agent}.
Most early benchmarks relied on final state matching or success rate (SR). However, researchers have recognized that SR often collapses into binary outcomes that fail to indicate specific failure modes. ProBench \citep{yang2025probench} addressed this by introducing a ``Process Provider" to capture accurate intermediate information for ``Process-related Tasks". Similarly, SPA-Bench \citep{chen2025spabench} and A3 \citep{chai2025a} explored step-level state validation but often relied on rigid predefined trajectories or potentially unreliable LLM-based decomposition. WindowsWorld improves upon these by integrating a more flexible checkpoint-based scoring system that allows for alternative valid paths while ensuring functional correctness at key process nodes.



\begin{figure*}[t]
    \centering
    \includegraphics[width=0.90\textwidth]{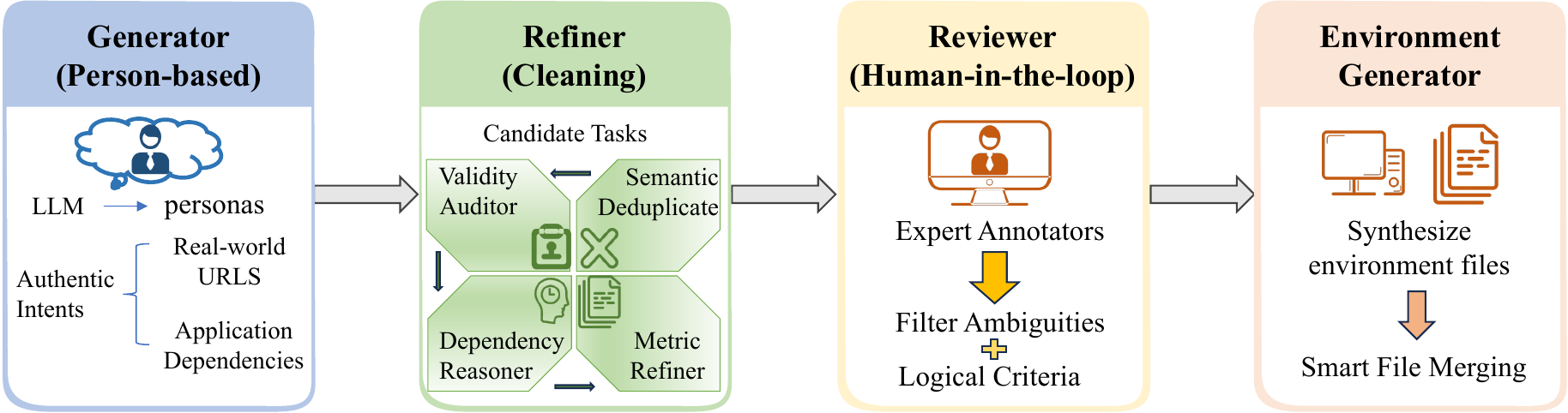} 
    \caption{The overall architecture of the human-in-the-loop multi-agent pipeline.}
    \label{fig:pipeline}
\end{figure*}

\section{WindowsWorld Benchmark}

WindowsWorld is a progress-centric benchmark for evaluating computer-use agents in realistic Windows desktop environments. All tasks are executed in controlled Windows virtual machines, where agents interact with a fixed set of commonly used productivity applications (see Figure~\ref{fig:table_left_image_right}) exclusively through standard GUI operations, without access to internal APIs or privileged system calls. This setup ensures reproducibility and ease of deployment.
In the following sections, we first give the task formulation of the computer-use agents (\S\ref{3.1}). We then describe the task and persona design that defines the benchmark's content space (\S\ref{3.2}). Next, we present a semi-automated task generation pipeline that combines LLM-based generation with systematic validation and human review (\S\ref{3.3}). We subsequently analyze the resulting benchmark in terms of task difficulty, cross-application complexity, and evaluation granularity (\S\ref{3.4}). Finally, we introduce the process-aware evaluation metrics (\S\ref{3.5}).

\subsection{Task Formulation}\label{3.1}
Following the standard protocol for autonomous agents, we formalize GUI interaction as a Partially Observable Markov Decision Process (POMDP) $(\mathcal{S}, \mathcal{O}, \mathcal{A}, \mathcal{T}, \mathcal{R})$ with state space $\mathcal{S}$, observation $\mathcal{O}$ (\S\ref{observation}), action space $\mathcal{A}$, transition function $\mathcal{T}: \mathcal{S} \times \mathcal{A} \rightarrow \mathcal{S}$, and reward function $\mathcal{T}: \mathcal{S} \times \mathcal{A} \rightarrow \mathbb{R}$.



At each step $t$, the agent executes a parameterized GUI action $a_t \in \mathcal{A}$ (e.g., mouse, keyboard, element-level operations) based on its observation, resulting in a new state $s_{t+1} \in \mathcal{S}$ and a new partial observation $o_{t+1}\in \mathcal{O}$. The agent repeats this process until output a termination mark (\texttt{DONE} or \texttt{FAIL}) or exceed maximum number of steps($t> t_{max}$), induces a trajectory
\begin{equation}
\tau = (o_1, a_1, \dots, o_T, a_T),
\end{equation}
The reward function $\mathcal{R}: \mathcal{S}\times \mathcal{A}\rightarrow [0,1]$ returns a non-zero value according to the query $q$ and the trajectory $\tau$ if the agent achieves the task objective, or accurately predicts failure for infeasible task.



\subsection{Task Taxonomy and Persona Design}
\label{3.2}

All tasks are organized along two complementary axes: task complexity and professional persona.

\textbf{Task Complexity Levels}. Tasks are categorized into four levels based on their cognitive and operational complexity:
L1 (\emph{Single-App Atomic}) tasks involve complex operations within a single application. 
L2 (\emph{Multi-App Linear}) tasks consist of sequential workflows that span multiple applications. 
L3 (\emph{Dynamic Reasoning}) tasks involve conditional logic or in‑process reasoning across multiple applications, e.g., tailoring actions based on whether certain information is present or absent.
L4 (\emph{Infeasible}) tasks are intentionally unachievable due to factors such as invalid URLs, missing files, or required account authentication.

\textbf{Persona Taxonomy}. To ensure that the generated instructions reflect authentic professional routines and heterogeneous workflows, we define a comprehensive taxonomy of 16 distinct personas. As illustrated in Figure~\ref{fig:table_left_image_right}, this taxonomy categorizes major professional domains and provides detailed persona configuration, capturing specific intents and constraints characteristic of real-world roles. By moving beyond generic commands, these personas ground our benchmark in diverse, domain-specific workflows.


\begin{figure*}[t]
    \centering
    \begin{subfigure}[b]{0.24\textwidth}
        \centering
        \includegraphics[width=\linewidth]{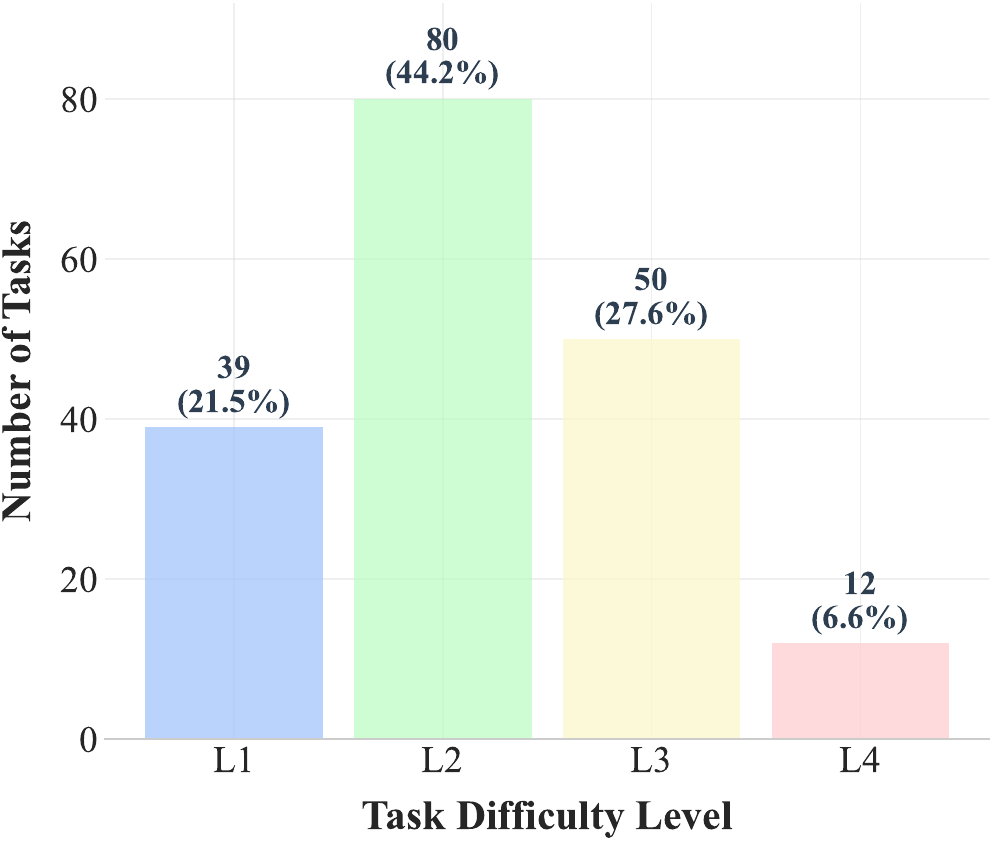}
        \caption{}
        \label{fig:detail_a}
    \end{subfigure}
    \hfill
    \begin{subfigure}[b]{0.24\textwidth}
        \centering
        \includegraphics[width=\linewidth]{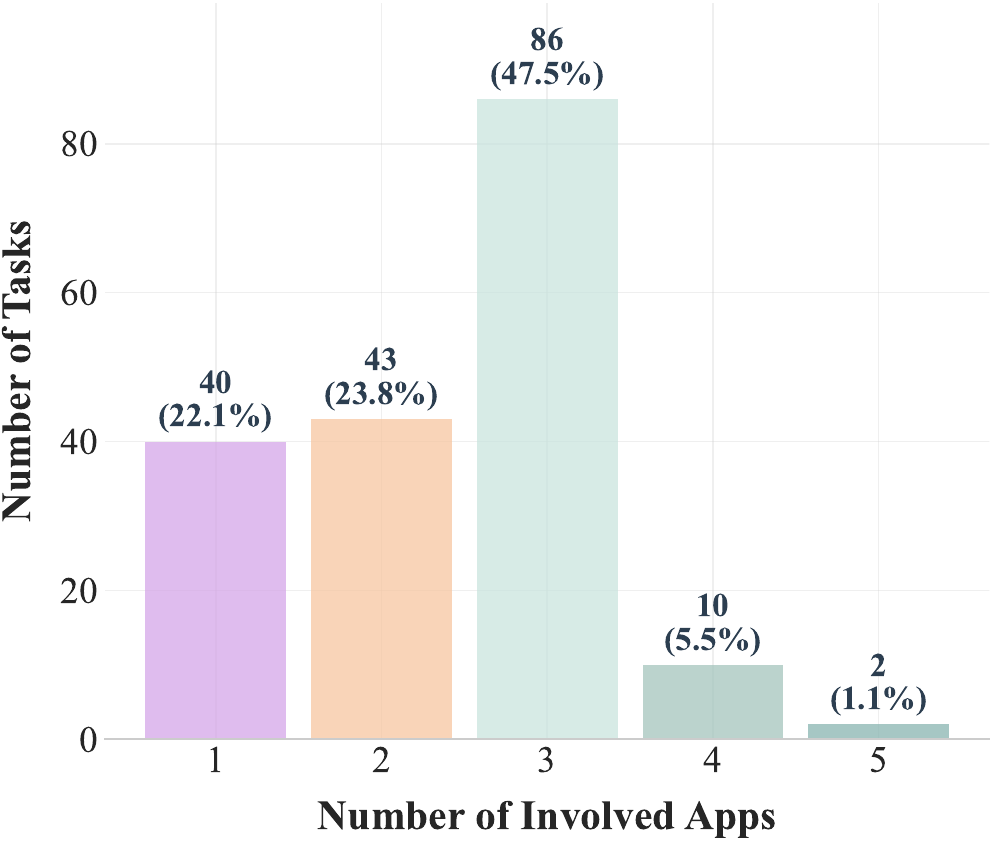}
        \caption{}
        \label{fig:detail_b}
    \end{subfigure}
    \hfill
    \begin{subfigure}[b]{0.24\textwidth}
        \centering
        \includegraphics[width=\linewidth]{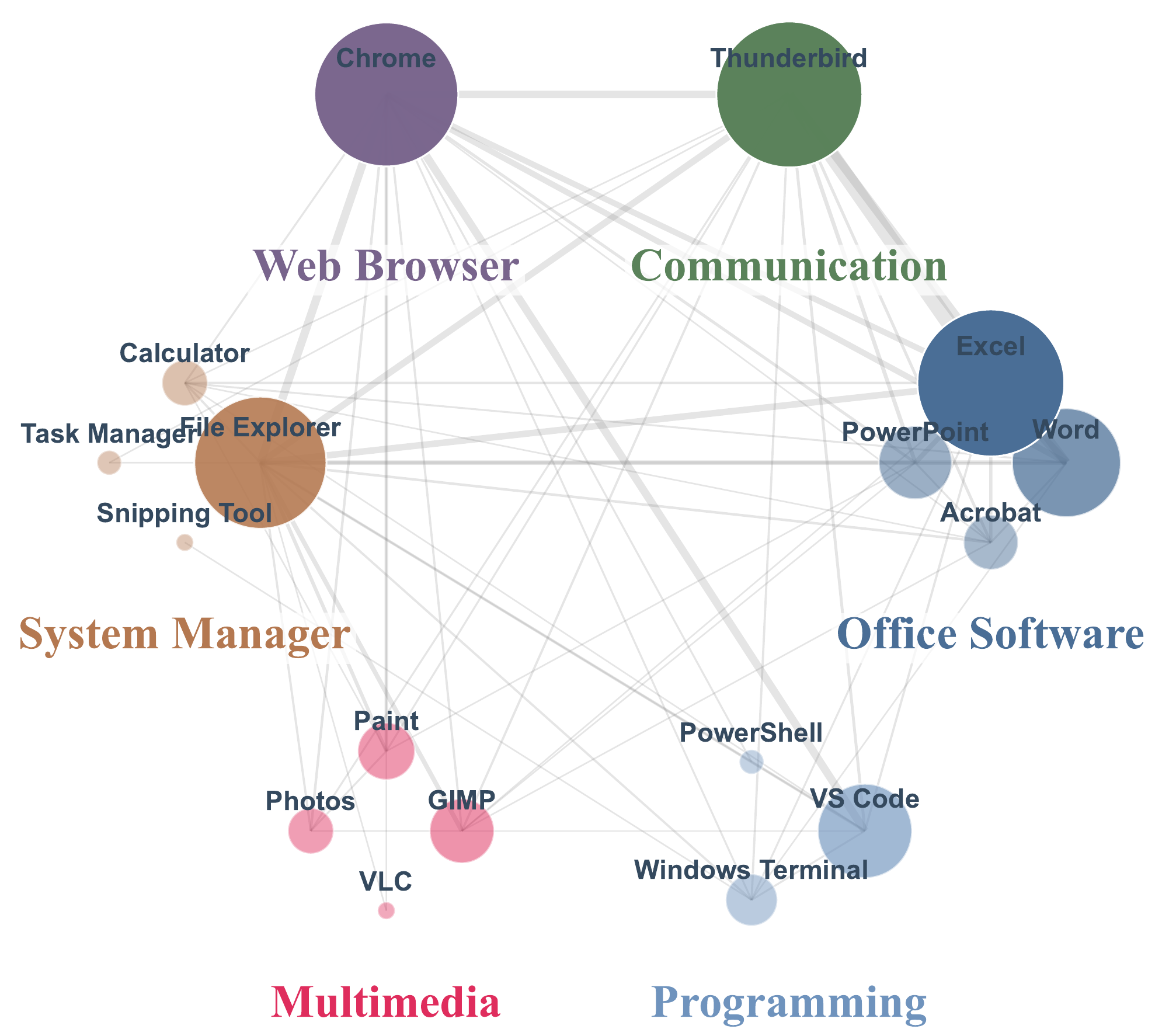}
        \caption{}
        \label{fig:detail_c}
    \end{subfigure}
    \hfill
    \begin{subfigure}[b]{0.24\textwidth}
        \centering
        \includegraphics[width=\linewidth]{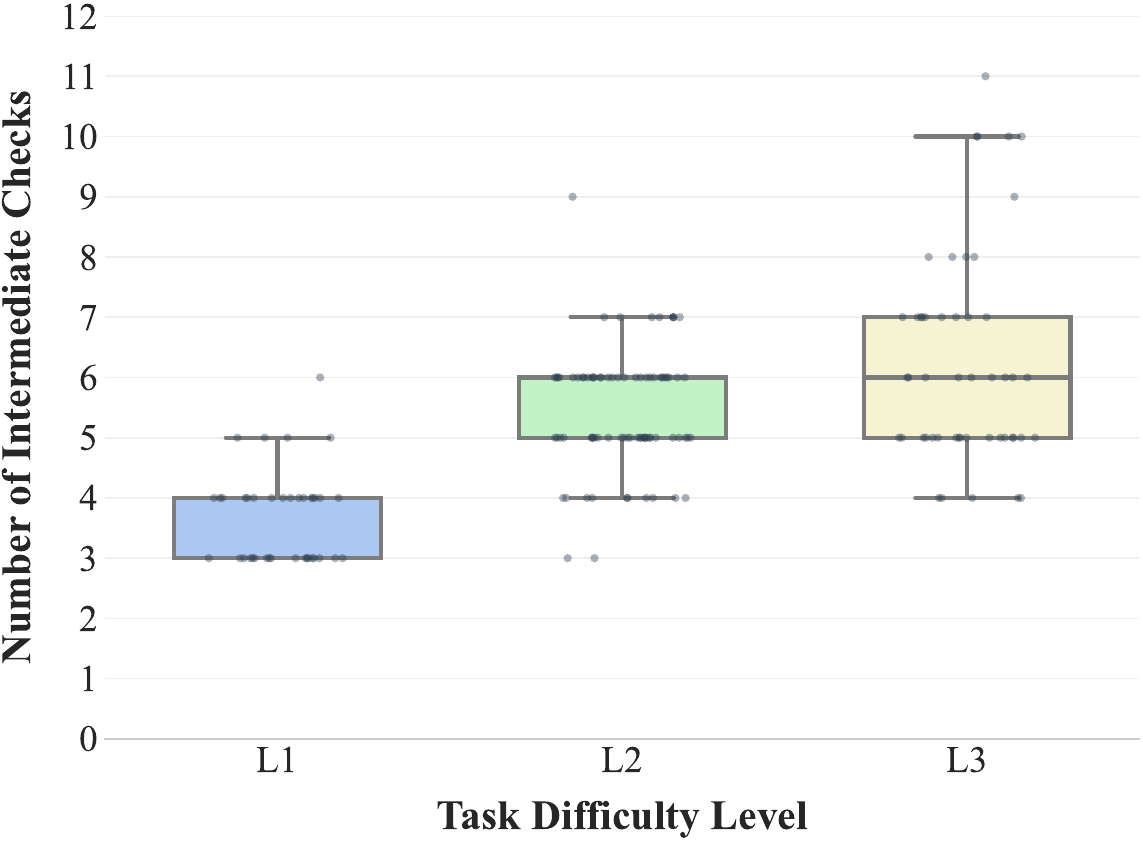}
        \caption{}
        \label{fig:detail_d}
    \end{subfigure}

    \caption{\textbf{Benchmark analysis of WindowsWorld}. 
(a) Distribution of tasks across difficulty levels (L1-L4), highlighting the prevalence of non-trivial multi-step workflows. 
(b) Distribution of the number of applications per task, highlighting the prevalence of multi-app workflows
(c) Cross-application interaction network, with edge weights representing co-occurrence frequency to reveal dependencies.
(d) Distribution of task checkpoints by difficulty (L1-L3), showing increased checkpoint density for complex tasks.}
\label{fig:combined_analysis}
\end{figure*}

\begin{table*}[t]
\centering
\small
\setlength{\tabcolsep}{5pt}
\renewcommand{\arraystretch}{1.0}
\begin{tabular}{lccccc}
\toprule
Benchmark & \#Apps & \#Tasks & Multi-app Tasks & Intermediate Checks & Platform \\


\midrule 

AndroidWorld \cite{rawles2024androidworld} & 20 & 116 & \ding{51}(9.50\%) & \ding{55} & Android \\
SPABench \cite{chen2025spabench} & 66 & 340 & \ding{51}(11.76\%) & \ding{55} & Android \\
ProBench \cite{yang2025probench} & 34 & 200 & -- & \ding{55} & Android \\

\midrule 

OSWorld (Windows) \citep{xie2024osworld} & 13 & 49 & \ding{51}(27.4\%)  & \ding{55} & Desktop  \\
Windows Agent Arena \citep{bonatti2024windows} & 11 & 154 & \ding{55} & \ding{55} & Desktop  \\
OSUniverse \citep{davydova2025osuniverse} & 9 & 160 & \ding{51}(26.8\%)  & \ding{55} & Desktop \\
WorldGUI \citep{zhao2025worldgui} & 10 & 611 & -- & \ding{55} & Desktop \\

\midrule 

WindowsWorld (ours) &17 & 181 & \ding{51}(77.9\%)  & \ding{51} (4.97/task) & Desktop \\
\bottomrule
\end{tabular}
\caption{\textbf{Comparison of execution-based benchmarks}. ``Multi-app'' indicates tasks with two or more applications; ``Intermediate Checks'' indicates tasks with intermediate-state checkpoints rather than result-only end-state evaluation. WindowsWorld contains the most apps and focuses on multi-app tasks across desktop benchmarks.}
\label{tab:benchmark-comparison}
\end{table*}

\subsection{Human-In-The-Loop Multi-Agent Framework}\label{3.3}
To reduce construction costs and speed up construction,
We propose a human-in-the-loop multi-agent pipeline to ensure ecological validity and scalability. Figure~\ref{fig:pipeline} provides an overview of the whole framework, where we describe each stage (agent) in detail below.

\begin{description}[style=unboxed, leftmargin=0pt, font=\bfseries]
    \item[Generator.] We leverage LLMs to generate tasks grounded in distinct professional personas (e.g., Accountant, Software Engineer). The generator receives structured prompts containing daily work routines and application-specific dependencies for each persona. In our implementation, we use DeepSeek-V3.2\cite{liu2025deepseek} as the generator and enable web search to retrieve and verify candidate URLs under an allowlist of open platforms. Unlike prior benchmarks relying on synthetic URLs, our generator is constrained to reference only accessible resources from open platforms (e.g., Github, Wikipedia, Stack overflow), ensuring that generated tasks reflect authentic web-integrated desktop workflows.
    \item[Refiner.] The candidate tasks undergo automated quality assurance through a 4-node pipeline: (i) \emph{Semantic Deduplicator}: Computes pairwise cosine similarity between task instruction embeddings and prunes near-duplicates exceeding a threshold ($\tau{=}0.85$), ensuring diversity across the dataset, (ii) \emph{Validity Auditor}: Performs asynchronous HTTP requests to verify URL accessibility and cross-references file mentions in instructions against the defined environment setup, flagging inconsistencies, (iii) \emph{Dependency Reasoner}: Employs LLM-based reasoning to transform procedural preconditions (e.g., "open the spreadsheet") into declarative environment states (e.g., "spreadsheet exists at specified path"), enabling deterministic setup verification, and (iv) \emph{Metric Refiner}: Reviews and standardizes evaluation criteria to ensure each intermediate checkpoint is unambiguous and verifiable against the expected task outcomes.
    \item[Human Reviewer.] Humans perform the final quality control to filter out tasks that automated validation cannot reliably detect.  We employ four annotators to check generated tasks, and they reject tasks of: (i) tasks with ambiguous or under-specified instructions that permit multiple valid interpretations, (ii) evaluation criteria that rely on subjective judgments rather than objective states, and (iii) tasks requiring unavailable proprietary software or inaccessible external services. 
    \item[Environment Generator.] For each validated task, the agent automatically synthesizes the required environment files(\texttt{.xlsx}, \texttt{.docx}, \texttt{.py}, etc.) using LLM-based content synthesis. We implement a Smart File Merging strategy that analyzes demands across multiple tasks within the same persona to create consistent data resources.
\end{description}

\subsection{Task Analysis}\label{3.4}

Using a human-in-the-loop framework, we construct WindowsWorld, a benchmark of 181 tasks across 17 Windows applications, organized into four difficulty levels (L1-L4). Beyond feasible tasks (L1-L3), we introduce infeasible L4 tasks to test an agent’s ability to reject unachievable goals. As shown in Figure~\ref{fig:detail_a} and \ref{fig:detail_b}, this benchmark emphasizes real-world multi-step workflows: 71.8\% of tasks are non-trivial (L2/L3), and 77.9\% involve two or more applications (avg. 2.4 apps/task). These multi-application workflows demonstrate structured dependencies (Figure~\ref{fig:detail_c}), such as recurring Excel-Thunderbird and Word-Thunderbird pairings that model common data-to-communication pipelines. To enable process-aware evaluation, tasks include explicit intermediate-state checkpoints (avg. 4.97 per task in Figure~\ref{fig:detail_d}) with higher-level tasks exhibiting denser checkpoints, thus enabling fine-grained diagnosis of long-horizon reasoning failures.




\textbf{Comparison with Existing Benchmarks}. As shown in Table~\ref{tab:benchmark-comparison}, WindowsWorld distinguishes itself through its exclusive focus on cross‑application coordination and professional workflows. While existing benchmarks like AndroidWorld and OSWorld remain largely confined to single‑app navigation (with multi‑app tasks comprising only 9.50\% and 27.4\% of their tasks, respectively), WindowsWorld is built around realistic multi‑app workflows, constituting 77.9\% of its evaluation suite. Moreover, unlike benchmarks that rely on \textit{binary ``all-or-nothing" scoring}, our benchmark introduces process-aware evaluation (avg. 4.97/task). This enables fine-grained assessment of partial progress, providing a diagnostic framework that reveals specific reasoning and execution bottlenecks in multi-step tasks.
To quantify horizon length beyond checkpoint counts, we estimate the average minimum number of expert actions required to complete each task. The average minimum action steps are 9.67 for L1, 18.13 for L2, and 27.81 for L3, confirming that WindowsWorld contains substantially longer-horizon workflows than prior Windows benchmarks.

\subsection{Process-Aware Evaluation}\label{3.5}

To address the lack of granularity in traditional ``all-or-nothing'' scoring for long-horizon GUI tasks, we propose a process-aware evaluation protocol. Formally, given an execution trajectory $\tau$, a set of intermediate checkpoints $\mathcal{C}=\{c_k\}$, and a terminal state $s_T$, we employ an automated judge $\mathcal{J}$ (implemented via Qwen3-VL-Plus) that outputs binary decisions $\mathcal{J}(\cdot) \in \{0,1\}$. The metrics are defined as follows:

\begin{equation}
\begin{aligned}
S_{\text{int}} &= \frac{1}{|\mathcal{C}|} \sum_{k=1}^{|\mathcal{C}|} \mathcal{J}(\tau, c_k), \\
S_{\text{final}} &= \mathcal{J}(\tau, s_T).
\end{aligned}
\label{eq:scores}
\end{equation}

\noindent \textbf{Intermediate Check Score ($S_{\text{int}}$):} Evaluated on L1-L3 tasks, this metric rewards partial progress by tracking essential sub-goals, preventing zero scores on complex tasks where a substantial portion of the workflow is completed.

\noindent \textbf{Checkpoint construction and path robustness.}
Intermediate checkpoints are first extracted by an LLM and refined by our Metric Refiner to remove action-specific constraints (e.g., ``click button X'') and retain only essential semantic states (e.g., ``the target file is open''). All checkpoints are then human-verified to be path-essential rather than trajectory-specific. Therefore, evaluation is state-based and tolerant to alternative valid execution paths, including shortcuts, menu actions, and keyboard commands.

\noindent \textbf{Reliability of the VLM judge.}
We evaluate $\mathcal{J}$ on 100 stratified tasks (L1: 24, L2: 50, L3: 26), covering 518 intermediate checkpoints. Two human annotators independently grade the trajectories, and we compare their consensus with the VLM judge. Agreement is strong: Pearson correlation is 0.9108 for $S_{\text{int}}$ and 0.8316 for $S_{\text{final}}$, while Cohen's $\kappa$ is 0.8668 at the checkpoint level and 0.8271 for final judgments. The corresponding 95\% confidence intervals are reported in Table~\ref{tab:judge_summary}, indicating near-human reliability. Representative false-positive and false-negative cases of the VLM judge are discussed in Appendix~\ref{app:vlm_judge}.


\noindent \textbf{Final Check Score ($S_{\text{final}}$):} Applied across L1-L4, this assesses the correctness of the terminal state, including the agent's ability to correctly reject infeasible instructions in Level-4.

\section{Experiment}
\subsection{Baselines}
We evaluate a diverse suite of SOTA agents to establish performance benchmarks on WindowsWorld, categorized into general-purpose multimodal models and specialized GUI agents. And we also set different input types for assessing the impacts of observation modalities. Details are shown in Appendix \ref{observation}.

\noindent \textbf{General-Purpose Models.} We select leading proprietary models to test foundational capabilities: Gemini-3-flash/pro-preview (\textit{20251217/20251118}), GPT-5.2 (\textit{20251211}), Claude-Sonnet-4.5 (\textit{20250929}), and Qwen3-VL-Plus \citep{Qwen3-VL}.

\noindent \textbf{GUI-Specialized Agents.} Complementing the general models, we evaluate domain-specific systems:
\textbf{Agent-S3} \citep{Agent-S3}: A research-oriented agent utilizing hierarchical task decomposition for complex interfaces.
\textbf{UiPath Screen Agent} \citep{UiPath}: An industrial-grade automation tool designed for enterprise screen-level interactions.




\begin{table*}[t]
\centering
\scriptsize
\setlength{\tabcolsep}{4pt}
\renewcommand{\arraystretch}{1.15}

\newcommand{\grouprow}[2]{%
  \specialrule{\lightrulewidth}{0pt}{0pt} 
  \rowcolor{#1}
  \multicolumn{10}{c}{\textit{#2}}\\ 
}

\label{tab:benchx-results}
\resizebox{\textwidth}{!}{%
\begin{tabular}{lccccccccc} 
\toprule
    & \multicolumn{4}{c}{$S_{\text{int}}$}
    & \multicolumn{5}{c}{$S_{\text{final}}$} \\
\cmidrule(lr){2-5}\cmidrule(lr){6-10} 
Model                 & L1      & L2      & L3      & Avg.    & L1      & L2      & L3      & L4      & Avg.    \\

\grouprow{catgreen}{Screenshot}
\hline

Gemini-3-pro-preview  & 46.20\% & 27.28\% & 24.88\% & 30.94\% & 25.64\% & 5.00 \% & 2.00 \% & 8.33 \% & 8.29 \% \\
Gemini-3-flash-preview       & 57.05\% & 45.73\% & 29.66\% & 43.58\% & 38.46\% & 15.00\% & 8.00 \% & 0.00 \% & 17.13\% \\
GPT-5.2               & 12.26\% & 6.56 \% & 5.51 \% & 7.69 \% & 5.13 \% & 1.33 \% & 0.00 \% & 0.00 \% & 1.78 \% \\
Claude-Sonnet 4.5     & 7.35 \% & 3.12 \% & 2.31 \% & 3.86 \% & 0.00 \% & 0.00 \% & 0.00 \% & 0.00 \% & 0.00 \% \\
Qwen3-vl-plus         & 26.37\% & 16.85\% & 10.69\% & 17.22\% & 2.56 \% & 0.00 \% & 0.00 \% & 0.00 \% & 0.55 \% \\

\grouprow{catblue}{Screenshot + Accessibility Tree (Hybrid)}
\hline

Gemini-3-pro-preview  & 63.33\% & 33.29\% & 29.68\% & 38.80\% & 41.94\% & 9.68 \% & 6.52 \% & 0.00 \% & 14.77\% \\
Gemini-3-flash-preview       & 67.52\% & 49.23\% & 38.63\% & 50.32\% & 35.90\% & 17.50\% & 14.00\% & 16.67\% & 20.44\% \\
GPT-5.2               & 13.63\% & 3.93 \% & 5.37 \% & 6.62 \% & 2.56 \% & 0.00 \% & 0.00 \% & 8.33 \% & 1.12 \% \\
Claude-Sonnet 4.5     & 16.20\% & 3.16 \% & 2.04 \% & 5.84 \% & 0.00 \% & 0.00 \% & 0.00 \% & 8.33 \% & 0.55 \% \\
Qwen3-vl-plus         & 22.52\% & 20.69\% & 14.79\% & 19.37\% & 7.69 \% & 1.25 \% & 10.00\% & 0.00 \% & 4.97 \% \\

\grouprow{catyellow}{Set of Marks}
\hline

Gemini-3-pro-preview  & 47.86\% & 33.13\% & 24.18\% & 33.88\% & 23.08\% & 7.50 \%  & 6.00\% & 8.33 \% & 9.94 \% \\
Gemini-3-flash-preview       & 58.85\% & 38.64\% & 25.38\% & 39.38\% & 30.77\% & 15.00\% & 2.00 \% & 0.00 \% & 13.81\% \\
GPT-5.2               & 3.42 \% & 3.61 \% & 1.25 \% & 4.57 \% & 0.00 \% & 1.25 \% & 0.00 \% & 25.00\% & 0.55 \% \\
Claude-Sonnet 4.5     & 12.05\% & 3.30 \% & 1.87 \% & 4.90 \% & 2.56 \% & 0.00 \% & 0.00 \% & 8.33 \% & 0.55 \% \\
Qwen3-vl-plus         & 6.62 \% & 2.12 \% & 5.88 \% & 4.27 \% & 5.13 \% & 0.00 \% & 4.00 \% & 0.00 \% & 2.21 \% \\

\grouprow{catred}{Agent}
\hline
UIPath w/ Gemini-3-flash-preview
                      & 21.67\% & 15.83\% & 7.34 \% & 14.96\% & 3.03 \% & 0.00 \% & 0.00 \% & 50.00\% & 4.64 \% \\
S3 w/ Qwen3-vl-plus   & 49.57\% & 30.31\% & 25.79\% & 33.47\% & 17.95\% & 3.75 \% & 2.00 \% & 16.67\% & 7.18 \% \\
S3 w/ Gemini-3-flash-preview
                      & 51.84\% & 39.28\% & 40.11\% & 42.42\% & 30.77\% & 16.25\% & 10.00\% & 8.33 \% & 17.13\% \\
\bottomrule

\end{tabular}%
}\caption{\textbf{Model and Agents Performance on WindowsWorld.}
All large models use a unified PyAutoGUI action space, while UiPath employs the \texttt{Computer\_13} action space from OSWorld.
Pure models are evaluated under Screenshot, Screenshot + Accessibility Tree, and Set-of-Mark inputs; Agent-based systems (S3 and UiPath) use Screenshot input. Moreover, S3 and UiPath are integrated UI-TARS-1.5-7B as a grounding model. Each task is executed under a fixed maximum step budget that depends on task level: \textbf{15 (L1), 25 (L2), 40 (L3), and 20 (L4)}. 
$S_{\mathrm{int}}$ averages L1--L3 intermediate checkpoints and $S_{\mathrm{final}}$ averages L1--L4 final task completion.}
\label{main_results}
\end{table*}




\subsection{Main Results and Analysis}

Table~\ref{main_results} presents the performance across all evaluated agents, revealing three critical insights into current GUI automation capabilities.

\noindent \textbf{The Efficiency-Completion Gap.} All models exhibit a steep performance decay as the task horizon and complexity increase. While the SOTA \textit{Gemini-3-flash} (Hybrid) maintains a reasonable intermediate progress score ($S_{\text{int}} = 50.32\%$), its success rate in reaching the terminal state drops to $S_{\text{final}} = 20.44\%$. This $30\%$ discrepancy suggests that agents frequently ``wander" through workflows—completing sub-goals but failing to synthesize them into a successful conclusion.

\noindent \textbf{Complexity Bottlenecks.} We observe a sharp decline in functional correctness as we transition from L1 to L3 tasks. For instance, \textit{Gemini-3-flash}'s $S_{\text{final}}$ plummets from $35.90\%$ (L1) to $16.67\%$ (L3). This underscores a fundamental weakness in cross-application coordination and state maintenance; agents excel at isolated operations but struggle with the conditional reasoning (L3) required for professional-grade workflows.

\noindent \textbf{Failure in Negative Constraint Handling (L4).} Current models lack the ``self-awareness" to identify infeasible instructions. Best \textit{GPT-5.2} (SoM) achieves only $25\%$ success rate on L4 tasks, indicating a high propensity for hallucinated success. UIPath performs poorly on general tasks, yet achieves a high L4 score because it often deems tasks incomplete.
The ability to reliably report task failure remains a significant open challenge.

\begin{figure}[t]
    \centering
    \begin{subfigure}[b]{0.48\columnwidth}
        \centering
        \includegraphics[height=5.0cm,keepaspectratio]{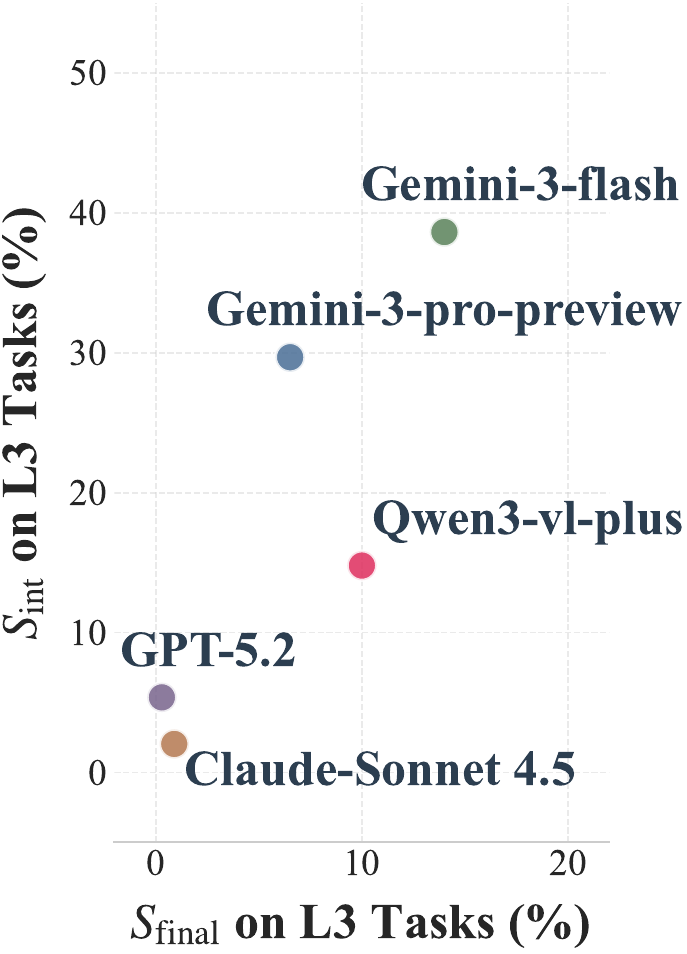}
        \caption{}
        \label{fig:intermediate_check_analysis}
    \end{subfigure}
    \hfill
    \begin{subfigure}[b]{0.48\columnwidth}
        \centering
        \includegraphics[height=5.0cm,keepaspectratio]{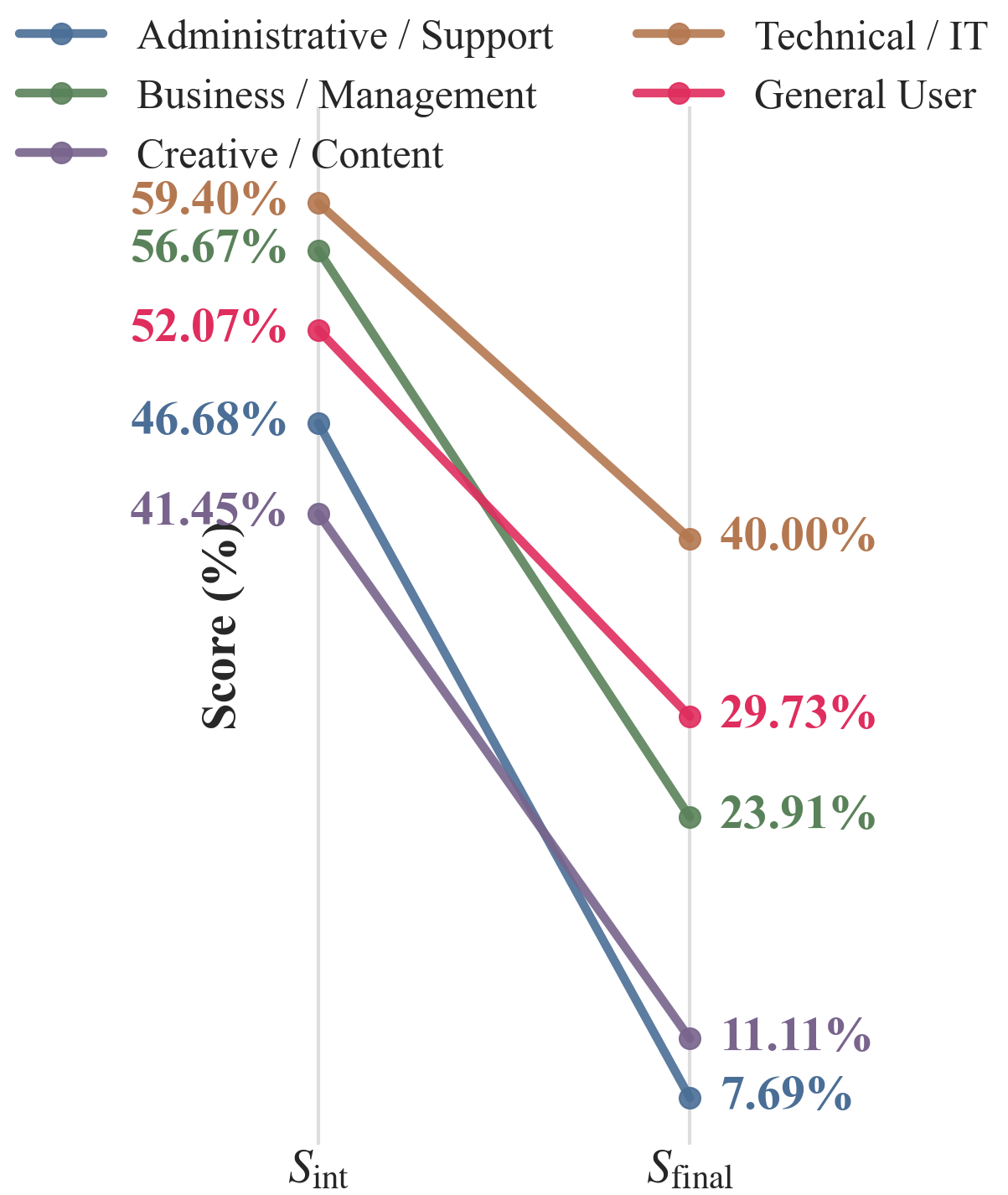}
        \caption{}
        \label{fig:persona_analysis}
    \end{subfigure}
    \caption{
    \textbf{Detailed analysis of intermediate checks and persona-dependent difficulty.}
    (a) Scatter plot of intermediate checkpoint score ($S_{\mathrm{int}}$) versus final task completion score ($S_{\mathrm{final}}$) on L3 tasks under the Screenshot + Accessibility setting. 
    (b) Model performance across different professional tasks.
    }
    \label{fig:detailed_analysis}
    \vspace{-0.5em}
\end{figure}


\subsection{In-Depth Analysis}

\noindent \textbf{Analysis of Observation Modalities.} 
Cross-modality evaluation identifies structured metadata (\textit{Hybrid}) as the most reliable scaffold, with \textit{Gemini-3-pro} leveraging accessibility trees to resolve UI density (gaining $+7.9\%$ $S_{\text{int}}$). In contrast, \textit{Raw Screenshot} reveals a grounding schism: \textit{Gemini-3-flash} excels in zero-shot coordinate mapping ($S_{\text{int}} = 43.5\%$), while \textit{Claude} and \textit{GPT-5.2} fail at pixel-level parsing ($<8\%$). Notably, \textit{Set-of-Marks} (SoM) yields inconsistent returns. It suggests that while heuristic visual overlays can aid localization, they often introduce cognitive noise that disrupts the internal spatial reasoning of VLMs, whereas structured metadata provides a more stable foundation for GUI models.

\noindent \textbf{Model Capabilities and Framework Synergy.} The evaluation reveals a performance hierarchy led by the \textit{Gemini-3-flash} series, which achieves a peak $S_{\text{int}}$ of 50.32\% under the \textit{Hybrid} modality. In contrast, \textit{GPT-5.2} exhibits a catastrophic failure in reasoning over structured interface data ($S_{final} \approx 0$), while \textit{Claude-Sonnet 4.5} and \textit{Qwen3-VL-Plus} show persistent gaps in GUI-specific visual reasoning. The integration of the \textit{S3} framework significantly boosts performance for grounding-weak models like \textit{Qwen3-VL-Plus} (nearly doubling $S_{\text{int}}$), yet offers diminishing returns for \textit{Gemini} models with strong native grounding. This suggests that external modules become less effective as the capability of the foundation model grows.

\noindent \textbf{Discriminative Power of Intermediate Checks.} As task complexity scales, end-state evaluation becomes increasingly uninformative due to low-score saturation. Figure~\ref{fig:intermediate_check_analysis} demonstrates this phenomenon on L3 tasks: while most models exhibit stagnant completion rates ($S_{\mathrm{final}} < 15\%$), their intermediate progress scores ($S_{\mathrm{int}}$) remain highly discriminative, ranging from $2.04\%$ to $38.63\%$. Even among agents with zero terminal success, $S_{\mathrm{int}}$ successfully distinguishes systematic progress from total failure. These results underscore that intermediate checks are essential for capturing latent competence in long-horizon tasks.


    

\noindent \textbf{Persona-Dependent Sensitivity.} Evaluation across personas (Figure~\ref{fig:persona_analysis}) shows stable intermediate progress but sharply diverging final success rates, highlighting a ``long-tail'' difficulty effect. Productivity-oriented personas exhibit a significant gap between $S_{\text{int}}$ and $S_{\text{final}}$, indicating that failures stem from late-stage coordination and global consistency rather than initial execution. Conversely, Technical/IT personas show tighter coupling between progress and completion. These disparities underscore that persona-specific difficulty in \textit{WindowsWorld} is driven by terminal-state complexity, validating intermediate checkpoints as critical diagnostic tools for identifying obscured execution bottlenecks (Table~\ref{tab:benchx-persona-results}).

\begin{table}[t]
\centering

\begin{subtable}[t]{\columnwidth}
\centering
\small
\setlength{\tabcolsep}{3pt}
\renewcommand{\arraystretch}{1.05}
\begin{tabular}{lccc}
\toprule
\textbf{Model} & \textbf{Shot} & \textbf{Shot+A11y} & \textbf{SoM} \\
\midrule
\multicolumn{4}{l}{\textit{Avg. latency (s/step)}} \\
Gemini-3-flash        & 10.60 & 9.60 & 16.95 \\ 
Gemini-3-pro-preview  & 19.32 & 23.83 & 43.69 \\
GPT-5.2               & 20.79 & 9.71 & 13.55 \\
Claude-Sonnet 4.5     & 11.55 & 24.97 & 25.84 \\
Qwen3-vl-plus         & 5.69  & 6.95  & 6.09 \\
\midrule
\multicolumn{4}{l}{\textit{Avg. steps (Gemini-3-flash): Failure$-$Success ($\Delta$)}} \\
L1 &  6.13 & 4.64  & 5.75 \\
L2 &  4.83 & 5.86  & 7.08 \\
L3 &  7.00 & 14.14 & 12.00 \\
L4 &  14.00 & 8.50 & 2.00 \\
\bottomrule
\end{tabular}
\caption{Efficiency under different input modalities.}
\label{tab:ablation-efficiency}
\end{subtable}

\vspace{0.6em} 

\begin{subtable}[t]{\columnwidth}
\centering
\small
\setlength{\tabcolsep}{5pt}
\renewcommand{\arraystretch}{1.05}
\begin{tabular}{lcccc}
\toprule
\textbf{Lang.} & $S_{\mathrm{int}}$ & $S_{\mathrm{final}}$ & $S_{\mathrm{final}}$ (L2) & $S_{\mathrm{final}}$ (L3) \\
\midrule
ZH & 43.59 & 17.13 & 15.00 & 8.00 \\
EN & 46.77 & 20.00 & 15.87 & 14.29 \\
\bottomrule
\end{tabular}
\caption{Instruction language.}
\label{tab:ablation-language}
\end{subtable}

\caption{\textbf{Ablation studies on efficiency and instruction language.} Latency is measured as the average per-step inference time. Step gap $\Delta$ denotes the difference between failed and successful trajectories (Failure$-$Success).}
\label{fig:ablation-study}
\vspace{-0.4em}
\end{table}

\begin{table}[t]
\centering
\scriptsize
\setlength{\tabcolsep}{3pt}
\renewcommand{\arraystretch}{1.05}
\resizebox{\linewidth}{!}{
\begin{tabular}{lcccccccccc}
\toprule
\textbf{Model} 
& $c_1$ & $c_2$ & $c_3$ & $c_4$ & $c_5$ 
& $c_6$ & $c_7$ & $c_8$ & $c_9$ & $c_{10}$ \\
\midrule
Gemini-3-pro-preview & 39.2 & 22.4 & 20.0 & 8.0 & 4.0 & 1.6 & 0.8 & 0.8 & 0.0 & 0.0 \\
Gemini-3-flash       
& 23.1 & 26.2 & 21.5 & 12.3 & 5.4 & 2.3 & 0.0 & 0.0 & 0.0 & 0.0 \\
S3 w/ Gemini-3-flash 
& 29.2 & 27.7 & 10.8 & 19.2 & 6.9 & 0.8 & 0.8 & 0.8 & 0.0 & 0.8 \\
S3 w/ Qwen3-vl-plus  
& 33.8 & 32.3 & 16.2 & 5.4 & 8.5 & 2.3 & 0.8 & 0.0 & 0.0 & 0.0 \\
\bottomrule
\end{tabular}}

\caption{\textbf{Checkpoint-wise First-Failure Distribution on WindowsWorld (Screenshot-only).} The percentage of trajectories failing first at each checkpoint $c_k$, evaluated on feasible multi-app tasks (L2--L3).}
\label{tab:checkpoint-failure-breakdown}
\end{table}

\noindent \textbf{Inference Efficiency and Failure Dynamics.} Table~\ref{tab:ablation-efficiency} highlights significant variance in per-step latency and execution patterns. While \textit{Qwen3-VL-Plus} maintains superior speed ($\sim$6s/step), \textit{Gemini-3-pro} exhibits high modality sensitivity, particularly under SoM. Notably, SoM-style inputs introduce non-trivial overhead even for efficient models like \textit{Gemini-3-flash}. Beyond latency, we observe that failed trajectories become increasingly inefficient as task difficulty scales. For \textit{Gemini-3-flash}, the failure-success step gap ($\Delta$) widens drastically from L1 to L3 (e.g., $4.64 \rightarrow 14.14$ in Hybrid), indicating that complex failures are rarely immediate but occur after extended, locally plausible execution. This ``inefficient drift" confirms that current agents struggle with global plan consistency and lack the self-correction to terminate efficiently when progress stalls.

\noindent \textbf{Impact of Instruction Language.} Cross-lingual evaluation (Table~\ref{tab:ablation-language}) reveals a persistent performance gap; English instructions consistently yield superior comprehension and planning stability over semantically equivalent Chinese prompts, particularly for \textit{Gemini-3-Flash}. This advantage is most pronounced in multi-step execution, where English better sustains trajectory consistency. These results highlight a linguistic bias in current SOTA agentic models.

\noindent \textbf{Error Analysis of Steps}. We presented the percentage of errors the model made under different sub-goals in Table \ref{tab:checkpoint-failure-breakdown}, finding that the model typically encountered errors at the beginning when dealing with cross-application tasks (L2--L3). Detailed case (in Appendix) studies revealed that the model performed poorly in Chinese file opening, cross-application information transfer.

\noindent \textbf{Cross-application difficulty beyond horizon length.}
A natural concern is whether multi-application tasks are simply harder because they are longer. To control for this factor, we compare a step-matched subset of L1 and L2 tasks with similar average minimum action steps (10.92 vs.\ 11.26). Despite comparable horizon length, performance drops sharply in the multi-application subset: $S_{\text{int}}$ decreases from 65.74\% to 35.14\%, and $S_{\text{final}}$ decreases from 46.15\% to 14.29\%. This suggests that context switching and cross-application state maintenance, rather than step count alone, are the dominant bottlenecks in WindowsWorld.

\section{Conclusion}
In this paper, we present the first process-aware computer-use benchmark in a cross-application environment. The challenging benchmarks include professional-grade multi-app tasks (78\%), fine-grained intermediate process checking, and an automatic task construction method. Experimental results show that the best agents only achieve a 20\% success rate.

\section*{Limitations}
Despite its benchmarking capabilities, our work has several limitations. \textbf{First}, our intermediate scores rely on full-trajectory execution and manually reviewed checkpoints, which restricts scalability for large-scale or online reinforcement learning (RL) in long-horizon tasks. Developing automated, generalizable reward formulations remains an open challenge. \textbf{Second}, while we evaluate cross-lingual instructions, the current environment is primarily optimized for a single-language OS interface; expanding to \textbf{multi-lingual OS environments} (e.g., non-English/Chinese UI elements) is necessary to assess true global robustness. \textbf{Finally}, \textit{WindowsWorld} does not yet incorporate evaluation for Model Context Protocol (MCP) tools, which we leave for future expansion.

\section*{Ethical Considerations and Reproducibility}
The data annotation and verification process involved four postgraduate researchers who filtered instructions and validated intermediate checkpoints. Annotators were compensated at a rate of 1.5 USD per task, reflecting fair market value for technical evaluation. As the instructions were generated through LLM-assisted pipelines and manually reviewed, they contain no personally identifiable information (PII) or harmful content. 

To foster community growth and ensure reproducibility, we commit to \textbf{open-sourcing our complete research framework}, including the \textit{WindowsWorld} evaluation environment, the full set of task instructions, and our process-aware evaluation metrics. All code and data will be made available on GitHub upon publication to support further research in GUI agents.

\section*{Acknowledgments}
This work was supported by the National Natural Science Foundation of China (Grant Nos. 62422603 and 625B2061), the Guangdong Basic and Applied Basic Research Foundation (Grant No. 2024B0101050003), and the Shenzhen Science and Technology Program (Grant No. ZDSYS20230626091203008).

\bibliography{custom}

\appendix

\section{Details of WindowsWorld Environment}

\subsection{Observation Modalities}\label{observation}

Diverging from prior works that rely on text-only (A11y) settings, we strictly focus on vision-centric modalities. This approach addresses the limitations of structural metadata, which is often incomplete or insufficient for visually dense professional interfaces \citep{koh2024visualwebarena}. We evaluate agents under three distinct settings:
1) \textbf{Raw Screenshot:} The agent operates solely on visual input, necessitating both semantic reasoning and precise coordinate grounding without auxiliary markers.
2) \textbf{Set-of-Marks (SoM):} Visual inputs are augmented with numbered bounding boxes overlaying interactive elements, reducing the grounding burden \cite{yang2023set}.
3) \textbf{Screenshot + Accessibility Tree (Hybrid):} A combination of visual data and structured metadata. We employ this modality primarily to assess the marginal utility of structural cues for open-source models.

\subsection{Action Space}

We provide two types of action spaces: free-form pyautogui and computer\_13 from OSWorld \cite{xie2024osworld}. We mainly use pyautogui action space in experiments. And detailed actions in computer\_13 are displayed in Table~\ref{tab:computer-actions}.


\begin{table}[t]
\centering
\small
\setlength{\tabcolsep}{8pt}
\renewcommand{\arraystretch}{1.15}
\begin{tabular}{ll}
\toprule
\textbf{Group} & \textbf{Action} \\
\midrule

\multirow{8}{*}{\texttt{computer.mouse}} 
& \texttt{MOVE\_TO()} \\
& \texttt{CLICK()} \\
& \texttt{MOUSE\_DOWN()} \\
& \texttt{MOUSE\_UP()} \\
& \texttt{RIGHT\_CLICK()} \\
& \texttt{DOUBLE\_CLICK()} \\
& \texttt{DRAG\_TO()} \\
& \texttt{SCROLL()} \\

\midrule

\multirow{5}{*}{\texttt{computer.keyboard}} 
& \texttt{TYPE()} \\
& \texttt{PRESS()} \\
& \texttt{KEY\_DOWN()} \\
& \texttt{KEY\_UP()} \\
& \texttt{HOTKEY()} \\

\midrule

\multirow{3}{*}{\texttt{system}} 
& \texttt{WAIT()} \\
& \texttt{FAIL()} \\
& \texttt{DONE()} \\

\bottomrule
\end{tabular}
\caption{\textbf{Action space in WindowsWorld.}
Actions are grouped into mouse control, keyboard input, and system-level control signals.}
\label{tab:computer-actions}
\vspace{-0.5em}
\end{table}

\section{Detailed Task Cases}
\label{app:case}

\subsection{Feasible Task}

\begin{figure*}[t]
    \centering
    \includegraphics[width=\textwidth]{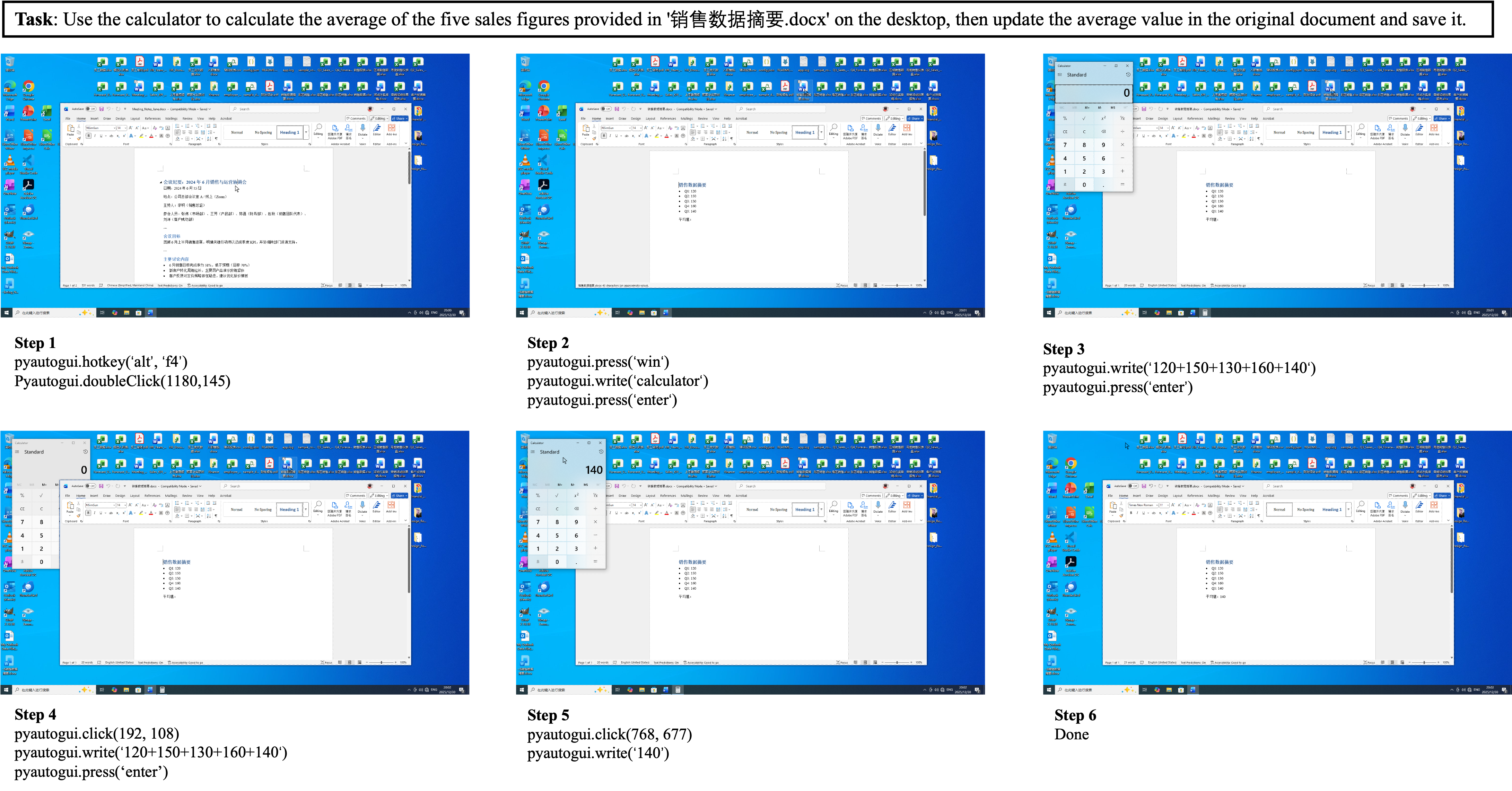}
    \caption{\textbf{Example execution trace on a feasible multi-app  WindowsWorld task (L1).}
    We visualize a representative successful trajectory, including the agent's observed GUI states and actions.
    Each panel corresponds to a key step in the trajectory, annotated with the executed operation.
    }
    \label{fig:feasible-run-visualization}
    \vspace{-0.8em}
\end{figure*}

For feasible tasks (L1--L3), we present example case in Figure~\ref{fig:l1-task-json-example}--\ref{fig:l3-task-json-example}, including the natural languange instruction, task category, apps involved the task and evaluation metrics for final criterion and intermediate checkpoints. In addition, we also provide visual cross-apps interactions sample for reference in Figure~\ref{fig:feasible-run-visualization}.

\begin{figure}[thp]
\centering
\begin{tcolorbox}[
    colback=gray!5, 
    colframe=gray!50, 
    left=2pt, right=2pt, top=2pt, bottom=2pt,
    boxrule=0.5pt,
    width=\columnwidth 
]
\begin{lstlisting}[
    language=json, 
    basicstyle=\ttfamily\scriptsize, 
    breaklines=true, 
    frame=none
]
{
  "instruction": "Open the file 'config.json' on the desktop in VS Code, use the built-in formatting tool to format the JSON content, and save the file.",
  "task_category": "L1",
  "involved_apps": [ "VS Code" ],
  "evaluation_metrics": {
    "success_criterion": "The config.json file has been correctly formatted and saved through the built-in formatting function of VS Code. The content is valid JSON with standard indentation.",
    "intermediate_checks": [
      "The config.json file has been successfully opened in VS Code",
      "The JSON content has been formatted using VS Code formatting commands (such as Shift+Alt+F)",
      "The file has been successfully saved and its content remains in a valid JSON structure"
    ]
  }
}
\end{lstlisting}
\end{tcolorbox}
\caption{Example WindowsWorld L1 task in JSON, including the natural-language instruction, involved applications, and intermediate/final evaluation criteria.}
\label{fig:l1-task-json-example}
\vspace{-0.5em}
\end{figure}

\begin{figure}[thp]
\centering
\begin{tcolorbox}[
    colback=gray!5, 
    colframe=gray!50, 
    left=2pt, right=2pt, top=2pt, bottom=2pt,
    boxrule=0.5pt,
    width=\columnwidth 
]
\begin{lstlisting}[
    language=json, 
    basicstyle=\ttfamily\scriptsize, 
    breaklines=true, 
    frame=none
]
{
  "instruction": "Open the downloaded image 'logo_raw.png' in the drawing tool, crop it to a square aspect ratio, then attach the cropped image and send an email to bench_serve1@2925.com with the subject 'Cropped Logo for Review'.",
  "task_category": "L2",
  "involved_apps": [ "Paint", "Thunderbird", "File Explorer" ],
  "evaluation_metrics": {
    "success_criterion": "Successfully sending the resized photo to email bench_serve1@2925.com and themed 'Cropped Logo for Review'.",
    "intermediate_checks": [
      "logo_raw.png is open",
      "The photo is resized to square",
      "The resized photo is saved as logo_cropped.png",
      "Added logo_cropped.png to an email and themed 'Cropped Logo for Review'"
    ]
  }
}
\end{lstlisting}
\end{tcolorbox}
\caption{Example WindowsWorld L2 task}
\label{fig:l2-task-json-example}
\vspace{-0.5em}
\end{figure}

\begin{figure}[thp]
\centering
\begin{tcolorbox}[
    colback=gray!5, 
    colframe=gray!50, 
    left=2pt, right=2pt, top=2pt, bottom=2pt,
    boxrule=0.5pt,
    width=\columnwidth 
]
\begin{lstlisting}[
    language=json, 
    basicstyle=\ttfamily\scriptsize, 
    breaklines=true, 
    frame=none
]
{
  "instruction": "Research 'obs-studio', 'shotcut', and 'gimp' on GitHub, visit their README pages, summarize their core features in a Word document, and include direct links to each repository. Save the document as 'Tool_Comparison.docx' on the desktop.",
  "task_category": "L3",
  "involved_apps": [ "Chrome", "Word", "File Explorer" ],
  "evaluation_metrics": {
    "success_criterion": "A Word document named 'Tool_Comparison.docx' exists on the desktop, containing the names of three GitHub content creation tools, a summary of their core features, and the corresponding full repository links.",
    "intermediate_checks": [
      "Chrome has opened the README pages of the three GitHub repositories.",
      "The key features of each tool have been extracted from the README and recorded in the Word document.",
      "The full URLs of each GitHub repository have been copied and included in the Word document.",
      "The Word document has been successfully saved as 'Tool_Comparison.docx' to the desktop."
    ]
  }
}
\end{lstlisting}
\end{tcolorbox}
\caption{Example WindowsWorld L3 task}
\label{fig:l3-task-json-example}
\vspace{-0.5em}
\end{figure}

\subsection{Infeasible Task}

For infeasible tasks (L4), we present example case in Figure~\ref{fig:l4-task-json-example}, including the same content in the feasible tasks case but except intermediate checkpoints.

\begin{figure}[thp]
\centering
\begin{tcolorbox}[
    colback=gray!5, 
    colframe=gray!50, 
    left=2pt, right=2pt, top=2pt, bottom=2pt,
    boxrule=0.5pt,
    width=\columnwidth 
]
\begin{lstlisting}[
    language=json, 
    basicstyle=\ttfamily\scriptsize, 
    breaklines=true, 
    frame=none
]
{
  "instruction": "Access https://gitee.com/nonexistent_user_2024/missing_project in Chrome, download the file 'config_template.json', and save it to the Downloads folder.",,
  "task_category": "L4",
  "involved_apps": [ "Chrome" ],
  "evaluation_metrics": {
    "success_criterion": "The Chrome TAB shows Gitee's 404 or the repository does not exist page. The Agent has output an error message indicating that the task is not feasible",
    "intermediate_checks": [ ],
  }
}
\end{lstlisting}
\end{tcolorbox}
\caption{Example WindowsWorld L4 task}
\label{fig:l4-task-json-example}
\vspace{-0.5em}
\end{figure}

\begin{table}[t]
\centering
\small
\setlength{\tabcolsep}{8pt} 
\renewcommand{\arraystretch}{1.4} 

\begin{tabularx}{\columnwidth}{@{}l>{\raggedright\arraybackslash}X@{}}
\toprule
\textbf{Level} & \textbf{Task Instruction} \\
\midrule

\multirow{2}{*}{L1}
& L1-1. Search for the official Chinese documentation of pandas in Chrome, look up \texttt{read\_excel}, copy the first paragraph into a new Notepad file \texttt{pandas\_read\_excel\_notes.txt}, and save it to the desktop. \\
\cmidrule(lr){2-2}
& L1-2. In File Explorer, navigate to \texttt{C:\textbackslash Logs}, create a folder \texttt{System\_Backup\_2024}, and set it to read-only. \\
\midrule

\multirow{2}{*}{L2}
& L2-1. Open \texttt{iris.csv} on the desktop, compute the average of \texttt{SepalLength}, write the result into a new VS Code file, and save it as \texttt{iris\_sepal\_avg.txt} on the desktop. \\
\cmidrule(lr){2-2}
& L2-2. Open the Quick Start Guide (\texttt{GitHub URL}), copy the “开发板” section, then compose an email in Thunderbird to \texttt{bench\_serve1@2925.com} with subject \texttt{[Support] OpenHarmony开发板要求} and include the copied content. \\
\midrule

\multirow{2}{*}{L3}
& L3-1. Open \texttt{Q2\_Financial\_Report\_Template.docx}, use \texttt{Q2\_Sales\_Data.xlsx} to compute the quarterly growth rate, fill the results, save as \texttt{Q2\_Financial\_Report\_Final.docx}, and email it to \texttt{bench\_serve1@2925.com}. \\
\cmidrule(lr){2-2}
& L3-2. Batch process PNGs in \texttt{Design\_Assets\_Q3} in GIMP (resize to $800{\times}600$, add a 2-pixel border, export JPEG at 85\%), then create \texttt{Q3\_Preview.pptx} in PowerPoint with title “Q3设计预览” and insert one processed sample image. \\
\midrule

\multirow{2}{*}{L4}
& L4-1. Download the latest \texttt{hosts} template from Gitee, save as \texttt{hosts\_template.txt} on the desktop, and open it in VS Code to verify the content. \\
\cmidrule(lr){2-2}
& L4-2. Access \url{https://gitee.com/nonexistent_user_2024/missing_project} in Chrome, download \texttt{config\_template.json}, and save it to Downloads. \\
\bottomrule
\end{tabularx}

\caption{\textbf{Task instances across difficulty levels (L1--L4).} Each level includes two representative instructions, ranging from single-app atomic operations (L1) to multi-app long-horizon workflows (L3) and infeasible tasks (L4).}
\label{tab:l1-l4-tasks}
\end{table}

\subsection{Error Case}

\begin{figure*}[t]
    \centering
    \includegraphics[width=\textwidth]{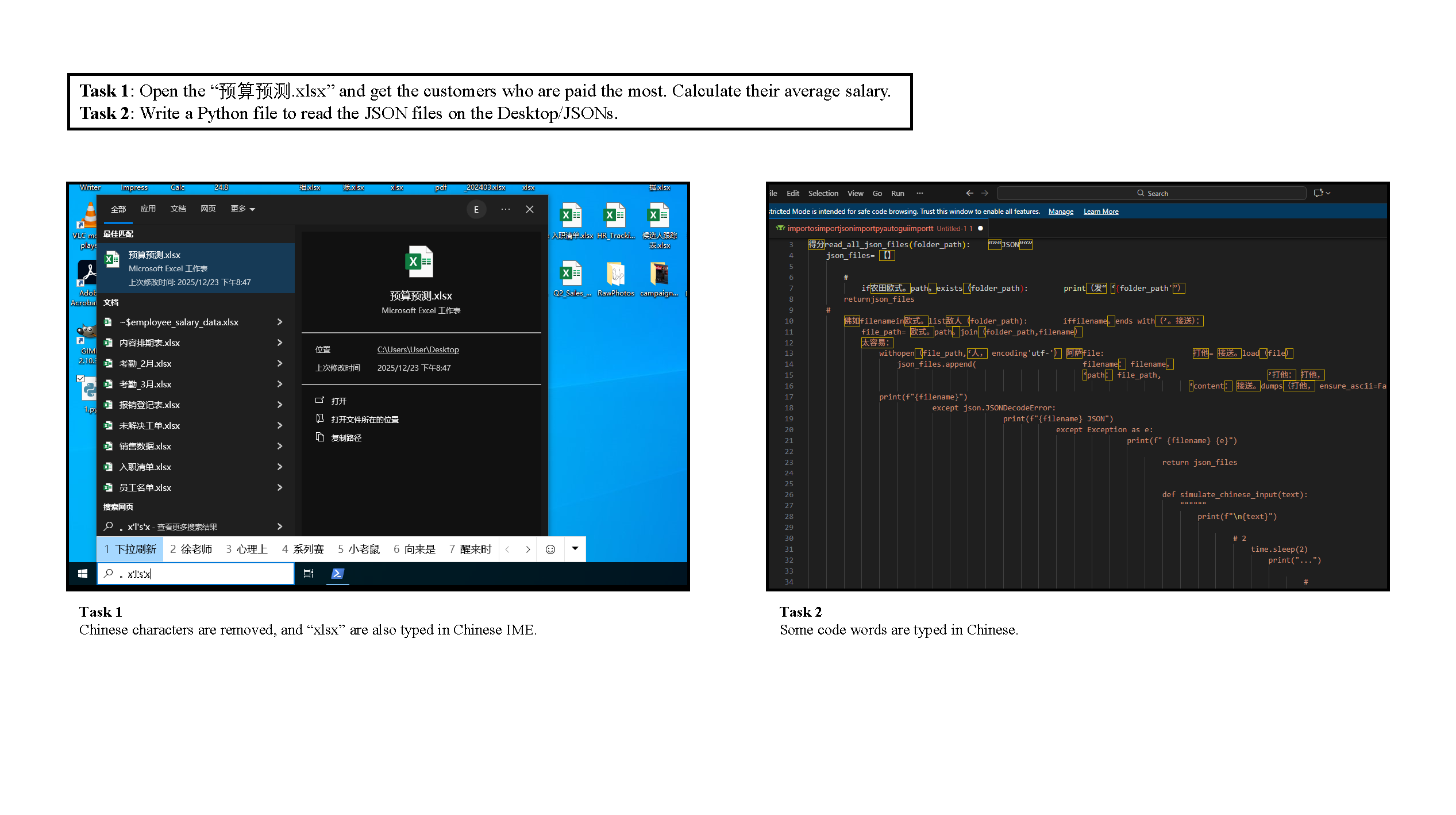}
    \caption{Technical challenges persist in the interaction between \texttt{pyautogui} and local Input Method Editors (IMEs), where Chinese input configurations often conflict with automated keystrokes. Furthermore, models frequently struggle to employ keyboard shortcuts for content synchronization across search interfaces. These issues highlight the need for agents to handle \textit{stochastic GUI anomalies}—such as IME pop-ups and inadvertent mouse triggers. Strengthening agent resilience to these environmental inconsistencies is vital for maintaining trajectory stability in complex desktop environments.
    }
    \label{fig:Chinese-conflict}
    \vspace{-0.8em}
\end{figure*}

The conflict between the \texttt{pyautogui} action space and Chinese keyboard settings is illustrated in Figure~\ref{fig:Chinese-conflict}. Since \texttt{pyautogui.write} simulates discrete keystrokes rather than direct string injection, non-ASCII characters are frequently intercepted by the system's Input Method Editor (IME). This causes the input to be misinterpreted as Pinyin, triggering unintended system shortcuts that lead to file-access errors and corrupted input strings. Such failures underscore the need for agents to possess cross-lingual environment awareness and robust text-entry strategies beyond basic keyboard simulation.

\begin{figure*}[t]
    \centering
    \includegraphics[width=\textwidth]{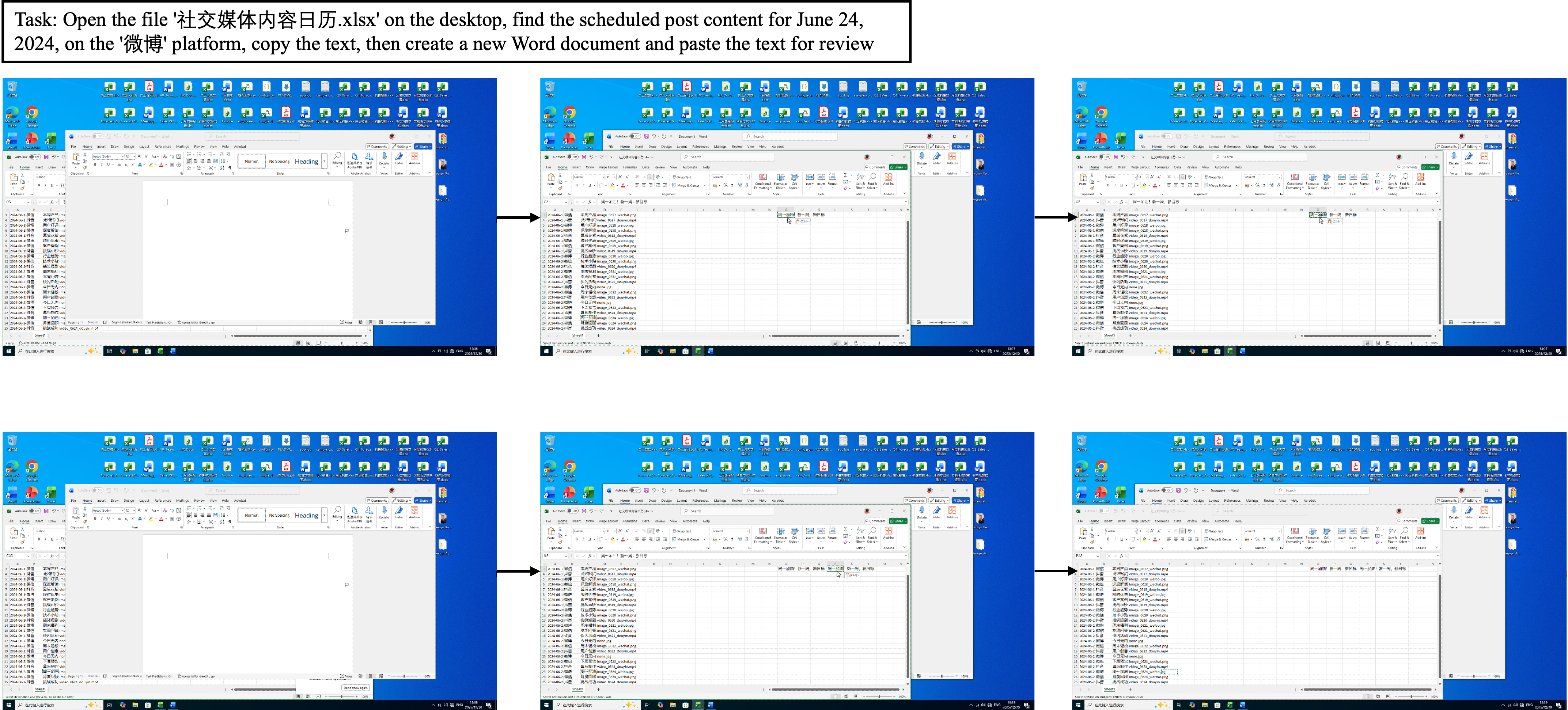}
    \caption{The model may suffer from the transfer between applications for cooperation (L2). However, an obstacle in application switching causes the agent to paste the content back into the source Excel spreadsheet. This persistent failure in synchronization is primarily driven by \textit{unstable window focus management}, where the model fails to confirm the active GUI context before executing clipboard actions.
    }
    \label{fig:l2-bad}
    \vspace{-0.8em}
\end{figure*}

We present Figure~\ref{fig:l2-bad} as a reference for failures in information transfer across software boundaries. In this scenario, the agent intends to migrate data from Excel to Word; however, an obstacle in application switching causes the agent to paste the content back into the source Excel spreadsheet. This persistent failure in synchronization is primarily driven by \textit{unstable window focus management}, where the model fails to confirm the active GUI context before executing clipboard actions. Such errors demonstrate that current agents lack the robustness required for multi-step, multi-application workflows, underscoring the need for explicit state-verification mechanisms during task transitions.

\section{Additional Experimental Results}

\subsection{Reliability and Error Analysis of the VLM Judge}\label{app:vlm_judge}

To validate the reliability of our automated VLM-as-judge protocol, we evaluate it on 100 stratified tasks (L1: 24, L2: 50, L3: 26), covering 518 intermediate checkpoints. Two human annotators manually annotate whether each intermediate checkpoint is satisfied from the execution trajectory, and we compare their consensus with the VLM judge (Qwen3-VL-Plus). As shown in Table~\ref{tab:judge_summary}, the agreement is strong: the Pearson correlations are 0.9108 for $S_{\mathrm{int}}$ and 0.8316 for $S_{\mathrm{final}}$, while Cohen's $\kappa$ reaches 0.8668 at the checkpoint level and 0.8271 for final judgments. These results support the reliability of VLM-based judging in our evaluation.


\begin{table}[t]
\centering
\small
\resizebox{\columnwidth}{!}{%
\begin{tabular}{lccc}
\toprule
Target & Metric & Value & 95\% CI \\
\midrule
$S_{\text{int}}$ & Pearson & 0.9108 & [0.8701, 0.9392] \\
$S_{\text{final}}$ & Pearson & 0.8316 & [0.7592, 0.8837] \\
Final judgment & Cohen's $\kappa$ & 0.8271 & [0.6687, 0.9544] \\
Checkpoint judgment & Cohen's $\kappa$ & 0.8668 & [0.8177, 0.9094] \\
\bottomrule
\end{tabular}%
}
\caption{Human validation of the VLM judge on 100 stratified tasks (518 checkpoints).}
\label{tab:judge_summary}
\end{table}

\noindent \textbf{Observed judge errors.}
Disagreements mainly arise from visual ambiguity rather than logical inconsistency. False positives occur when an action appears successful in the trajectory history but the actual GUI state is occluded by window changes. False negatives occur when the target state is briefly achieved but becomes invisible in the screenshot (e.g., after minimizing or switching windows).

\subsection{Persona-Wise Results}\label{app-persona}

Table~\ref{tab:benchx-persona-results} reports performance stratified by five persona categories, evaluated with both intermediate checkpoint score ($S_{\mathrm{int}}$ and final completion score ($S_{\mathrm{final}}$) under different input modalities and agent frameworks. This breakdown reveals substantial heterogeneity in difficulty across professional workflows, which is obscured by aggregate scores.

\paragraph{Progress--completion gap across personas.}
Across personas, a recurring pattern is the presence of large progress--completion gaps: agents often achieve non-trivial intermediate progress while failing to reach correct terminal states. This indicates that many failures occur after partial task execution rather than at the initial interaction stage.

\paragraph{Productivity-oriented personas.}
This gap is particularly pronounced in productivity-oriented personas (e.g., Administrative/Support and Creative/Content), where tasks are document-centric or visually driven and require maintaining global consistency across multiple steps. Although agents frequently succeed in local operations, they struggle with late-stage coordination and constraint satisfaction, leading to low final completion rates.

\paragraph{Structured Technical/IT workflows.}
In contrast, Technical/IT personas exhibit a tighter coupling between intermediate progress and final success. Tasks dominated by code editing and command-line interactions feature explicit goals and deterministic state transitions, under which current models perform comparatively better.

\paragraph{Implications for evaluation.}
These persona-wise differences suggest that difficulty in WindowsWorld is not uniform and that failures frequently arise in late-stage coordination rather than early execution. Consequently, intermediate checkpoint scores provide essential diagnostic signal for understanding model behavior across heterogeneous professional workflows, complementing final-state evaluation.



\begin{table*}[thp]
\centering
\scriptsize
\setlength{\tabcolsep}{4pt}
\renewcommand{\arraystretch}{1.15}

\newcommand{\grouprow}[1]{
    \rowcolor{gray!15}
    \multicolumn{11}{c}{\textit{#1}}
}

\newcommand{\inter}{$S_{\text{int}}$}
\newcommand{\final}{$S_{\text{final}}$}

\resizebox{\textwidth}{!}{
\begin{tabular}{lcccccccccc}
    \toprule
        & \multicolumn{2}{c}{Administrative / Support} 
        & \multicolumn{2}{c}{Business / Management}
        & \multicolumn{2}{c}{Creative / Content}
        & \multicolumn{2}{c}{Technical / IT}
        & \multicolumn{2}{c}{General User} \\
Model                  & \inter  & \final  & \inter  & \final  & \inter  & \final  & \inter  & \final  & \inter  & \final  \\
\midrule

\grouprow{Screenshot} \\
Gemini-3-pro-preview   & 29.34\% & 0.00 \% & 30.88\% & 8.70 \% & 23.68\% & 4.40 \% & 37.27\% & 20.00\% & 38.38\% & 16.22\% \\
Gemini-3-flash-preview & 35.92\% & 15.38\% & 43.38\% & 10.87\% & 36.77\% & 8.89 \% & 54.95\% & 33.33\% & 52.93\% & 35.14\% \\
GPT-5.2                & 5.00 \% & 0.00 \% & 6.19 \% & 0.00 \% & 4.35 \% & 0.00 \% & 7.78 \% & 6.67 \% & 15.49\% & 5.88 \% \\
Claude-Sonnet 4.5      & 4.01 \% & 0.00 \% & 4.03 \% & 0.00 \% & 3.44 \% & 0.00 \% & 6.89 \% & 0.00 \% & 11.85\% & 0.00 \% \\
Qwen3-vl-plus          & 21.56\% & 0.00 \% & 20.69\% & 0.00 \% & 12.73\% & 0.00 \% & 19.27\% & 0.00 \% & 14.50\% & 2.70 \% \\
\midrule

\grouprow{Screenshot + Accessibility Tree} \\
Gemini-3-pro-preview   & 26.26\% & 7.69 \% & 24.70\% & 2.17 \% & 27.58\% & 15.15\% & 31.24\% & 20.00\% & 42.66\% & 18.92\% \\
Gemini-3-flash-preview & 46.68\% & 7.69 \% & 56.67\% & 23.91\% & 41.45\% & 11.11\% & 59.40\% & 40.00\% & 52.07\% & 29.73\% \\
GPT-5.2                & 5.69 \% & 0.00 \% & 6.43 \% & 2.17 \% & 3.61 \% & 0.00 \% & 5.40 \% & 0.00 \% & 11.40\% & 0.00 \% \\
Claude-Sonnet 4.5      & 0.77 \% & 0.00 \% & 3.67 \% & 0.00 \% & 2.44 \% & 0.00 \% & 3.33 \% & 0.00 \% & 8.20 \% & 0.00 \% \\
Qwen3-vl-plus          & 19.44\% & 3.85 \% & 20.38\% & 2.17 \% & 15.59\% & 2.22 \% & 12.22\% & 6.67 \% & 25.54\% & 13.51\% \\
\midrule

\grouprow{Set of Marks} \\
Gemini-3-pro-preview   & 34.26\% & 7.69 \% & 30.96\% & 4.35 \% & 23.89\% & 6.67 \% & 47.78\% & 33.33\% & 43.78\% & 16.22\% \\
Gemini-3-flash-preview & 39.33\% & 7.69 \% & 41.53\% & 17.39\% & 29.26\% & 4.55 \% & 36.95\% & 20.00\% & 50.81\% & 27.03\% \\
GPT-5.2                & 0.00 \% & 0.00 \% & 3.79 \% & 0.00 \% & 4.07 \% & 0.00 \% & 1.90 \% & 0.00 \% & 10.45\% & 2.70 \% \\
Claude-Sonnet 4.5      & 2.82 \% & 0.00 \% & 3.18 \% & 0.00 \% & 5.81 \% & 0.00 \% & 0.00 \% & 0.00 \% & 9.37 \% & 2.70 \% \\
Qwen3-vl-plus          & 3.08 \% & 3.85 \% & 3.39 \% & 2.17 \% & 1.74 \% & 0.00 \% & 5.41 \% & 0.00 \% & 8.92 \% & 5.41 \% \\
\midrule

\grouprow{Agent} \\
UIPath w/ Qwen3-vl-plus
                       & 30.28\% & 0.00 \% & 33.86\% & 0.00 \% & 31.46\% & 4.44 \% & 36.06\% & 20.00\% & 36.62\% & 16.22\% \\
UIPath w/ Gemini-3-flash-preview
                       & 33.32\% & 15.38\% & 39.09\% & 13.04\% & 43.54\% & 15.56\% & 54.60\% & 33.33\% & 46.67\% & 21.62\% \\
S3 w/ Gemini-3-flash-preview
                       & 18.01\% & 0.00 \% & 19.48\% & 2.86 \% & 13.90\% & 0.00 \% & 6.06 \% & 0.00 \% & 12.56\% & 0.00 \% \\
\bottomrule

\end{tabular}
}
\caption{Main results on WindowsWorld cross input modalities and agent frameworks. Reported by persona category.}
\label{tab:benchx-persona-results}
\end{table*}

\section{Detailed multi-agent framework}

\subsection{Prompt for Generator}

In Figure~\ref{fig:prompt-generator}, we present the system prompt used in Generator agent in our human-in-the-loop multi-agent framework. This system prompt is input into the DeepSeek v3.2 model, along with various persona settings.

\begin{figure*}[p]
\centering
\promptbox{
You are an expert Data Engineer specializing in building benchmarks for GUI Agents. Your task is to generate realistic, complex, and structured user commands for a Windows-based Agent.

\textbf{\# Current Generation Context}
- \textbf{Target Persona:} \{persona\}
- \textbf{Task Category:} \{category\} (\{category\_desc\})
- \textbf{Number of Tasks to Generate:} \{batch\_size\}

\textbf{\# Persona Details}
The \{persona\} typically uses these applications:
- \textbf{Primary Apps:} \{primary\_apps\}
- \textbf{Secondary Apps:} \{secondary\_apps\}
- \textbf{Typical Work Tasks:} \{typical\_tasks\}

\textbf{\# Task Categorization Rules}
- \textbf{L1 (Single-App Atomic):} Operations within a SINGLE application only
- \textbf{L2 (Multi-App Linear):} Involves 2-3 apps with clear sequential flow
- \textbf{L3 (Dynamic Planning \& Reasoning):} Involves 3+ apps OR requires conditional logic/math/decision making
- \textbf{L4 (Impossible/Negative Tasks):} Tasks that CANNOT be completed (missing files, dead URLs, etc.)

\textbf{\# Output Format}
You MUST output a JSON array containing exactly \{batch\_size\} task objects. Each object must follow this schema:

\{
  "task\_id": "win\_\{persona\_id\}\_\{category\}\_XXX",
  "instruction": "Natural language command (in English)",
  "instruction\_cn": "Natural language command (in Chinese)",
  "task\_category": "\{category\}",
  "persona": "\{persona\}",
  "involved\_apps": ["MainApp", "SecondaryApp1", "SecondaryApp2"],
  "pre\_conditions": [
    "employee\_data.xlsx exists on Desktop",
    "Outlook is logged into company email"
  ],
  "environment\_setup": \{
    "files\_to\_create": [...],
    "urls\_to\_mock": [...],
    "downloadable\_resources": [...]
  \},
  "evaluation\_metrics": \{
    "success\_state": "Description of the final state when task is fully successful",
    "intermediate\_checks": [...]
  \}
\}

\textbf{\# CRITICAL: environment\_setup Usage Rules}

\textbf{environment\_setup is NOT a required field!} Only populate it when the task genuinely requires pre-existing files.

\textbf{When environment\_setup IS needed:}
- Task instruction explicitly mentions "open a file", "process a document", "edit a spreadsheet"
- Task requires operating on an \textbf{already existing} local file
- Example: "Open employee\_data.xlsx on Desktop, filter Engineering department employees"

\textbf{When environment\_setup is NOT needed (leave empty):}
- Task involves searching/browsing web content (e.g., "Search xxx in Chrome")
- Task involves creating new files (e.g., "Create a new Word document")
- Task involves sending emails (without attachments)
- Task involves downloading content from websites then processing
- Task uses built-in system features (Calculator, Paint for new drawings)

\textbf{\# Intermediate Checks - CRITICAL Guidelines}

- intermediate\_checks MUST list ALL MANDATORY steps required to complete the task
- Each check represents a REQUIRED milestone that CANNOT be skipped
- List checks in chronological order matching the task flow
- Use observable, verifiable states (e.g., "File saved", "Email sent")
- Do NOT include optional steps or alternative paths
- For LLM-based verification - describe states, NOT assertions or code

\textbf{MOST IMPORTANT: NEVER FABRICATE SPECIFIC DETAILS!}
The check content MUST match the EXACT specificity level of the instruction.

\textbf{Rule: If specific value is NOT in instruction, it should NOT appear in check!}

\textbf{\# CONSISTENCY RULE - Placeholders \& Specific Values Must Match Across All Fields}

- \textbf{instruction}, \textbf{instruction\_cn}, \textbf{intermediate\_checks}, and \textbf{success\_criterion} MUST use IDENTICAL placeholders/values

If instruction says "filter by 'Engineering' department":
  - intermediate\_check: "Filtered 'Engineering' department candidates"
  - success\_criterion: "Document contains Engineering department candidate information"

\textbf{The same placeholder in instruction should appear in checks exactly as specified.}

\textbf{\# Category-Specific Rules for L1 (Single-App Atomic) Tasks}

- Focus on single application operations
- Examples: Open a file, format a document, send an email, run a script
- Only ONE app should be in involved\_apps (this is automatically the primary app)
- Pre-conditions should relate to that single app being ready
- \textbf{Intermediate Checks:} List 2-4 MANDATORY steps 
  (e.g., "File opened" -> "Content modified" -> "File saved")

\textbf{CRITICAL - DO NOT fabricate specific values in checks:}
- WRONG: "Email subject is 'Interview Invitation - John Smith'" (fabricated name)
- CORRECT: "Email subject is 'Interview Invitation - [Candidate Name]'" (matches instruction placeholder)
- WRONG: "Email body contains interview time (July 10, 2024 at 10AM, Teams meeting)" (fabricated details)
- CORRECT: "Email body contains standard interview information (time, location, required materials)" (generic description)
- If instruction says "schedule details" without specific date, check should say "Schedule filled" NOT a made-up date

\textbf{\# Category-Specific Rules for L2 (Multi-App Linear) Tasks}

- Involve exactly 2-3 applications
- Clear sequential flow (MainApp -> SecondaryApp -> ...)
- \textbf{IMPORTANT:} List apps in order of PRIMARY usage. The first app should be where MOST work happens.
- Examples: ["Excel", "Word"] for "Copy Excel data and format in Word" (Excel is primary)
- Each step should naturally lead to the next
- \textbf{Intermediate Checks:} List 3-6 MANDATORY steps covering ALL app transitions 
  (e.g., "Excel data extracted" -> "Word document created" -> "Data pasted" -> "Document saved")

\textbf{CRITICAL - Checks must match instruction specificity level:}
- If instruction says "copy the data", check should say "Data copied" NOT list specific data content
- If instruction says "extract names", check should say "Names extracted" NOT "Extracted John, Mary, Tom"
- If instruction gives specific values (e.g., "filter by 'Engineering' department"), check CAN include "Filtered 'Engineering'"
- NEVER invent dates, names, numbers, or content not explicitly stated in instruction

\textbf{\# Category-Specific Rules for L3 (Dynamic Planning \& Reasoning) Tasks}

- Involve 2+ applications OR complex reasoning
- \textbf{IMPORTANT:} The FIRST app in involved\_apps should be the PRIMARY app where most complex work happens
- Include conditional logic: "if...then...", "based on...determine..."
- Include calculations or data aggregation across sources
- Examples: ["Excel", "Outlook", "Chrome"] for "Analyze Excel data, research online, send email summary" (Excel is primary)
- \textbf{Intermediate Checks:} List 4-8 MANDATORY steps including decision points 
  (e.g., "Data loaded" -> "Condition evaluated" -> "Calculation completed" -> "Results summarized" -> "Email sent")

\textbf{CRITICAL - Keep checks abstract for dynamic content:}
- For conditional tasks: "Condition evaluated" NOT "Evaluation result is yes" (result depends on actual data)
- For calculations: "Average calculated" NOT "Average is 85000" (value depends on data)
- For extracted content: "Key information extracted" NOT specific extracted text
- Checks should be verifiable regardless of actual data values

\textbf{\# Category-Specific Rules for L4 (Impossible/Negative) Tasks}

- Design tasks that WILL FAIL by design
- Examples: 
  - Open non-existent file "missing\_report\_2024.xlsx"
  - Navigate to broken URL "https://internal.company.fake/dashboard"
  - Send email to non-existent contact
- The goal is to test if the agent can properly report errors and handle failure cases
- Include clearly invalid resources in environment\_setup

}
\caption{System prompt used for persona-conditioned task generation in WindowsWorld. Placeholders (e.g., \{persona\}, \{primary\_apps\}) are instantiated during generation.}
\label{fig:prompt-generator}
\end{figure*}

\subsection{Prompt for Refiner}

In Figure~\ref{fig:prompt-dependency-reasoner}--\ref{fig:prompt-metric-refiner}, we present the system prompt used in Refiner agent in our human-in-the-loop multi-agent framework.

\begin{figure*}[p]
\centering
\promptbox{
\textbf{\# Dependency Reasoner - Converting Actions to States}

You are a professional task analysis expert. Please convert the following "action-form" pre-conditions into "state-form".

\textbf{Original Task Instruction:} \{instruction\}

\textbf{Current Pre-conditions (may be in action form):}
\{pre\_conditions\}

\textbf{Conversion Rules:}
\begin{itemize}
\item "Open Excel" $\rightarrow$ "Excel application is open and active"
\item "Log into email" $\rightarrow$ "User is logged into Outlook/email account"
\item "Connect to network" $\rightarrow$ "Network connection is stable and available"
\item "Open file X" $\rightarrow$ "File X exists at the specified path and is accessible"
\end{itemize}

\textbf{Output Requirements:}
\begin{itemize}
\item Output in JSON array format
\item Each condition should describe a specific state
\item Include necessary environment states (network, login status, etc.)
\item Output ONLY the JSON array, no other text
\end{itemize}
}
\caption{Prompt for the Dependency Reasoner node in the refiner pipeline. This module converts action-based pre-conditions into verifiable state descriptions.}
\label{fig:prompt-dependency-reasoner}
\end{figure*}

\begin{figure*}[p]
\centering
\promptbox{
\textbf{\# Metric Refiner - Generating Programmable Evaluation Assertions}

You are a professional test engineer. Please generate programmable evaluation assertions for the following GUI Agent task.

\textbf{Task ID:} \{task\_id\}
\textbf{Task Instruction:} \{instruction\}
\textbf{Involved Apps:} \{involved\_apps\}
\textbf{Task Category:} \{task\_category\}

\textbf{Current Evaluation Metrics:}
\begin{itemize}
\item Success Criterion: \{success\_criterion\}
\item Intermediate Checks: \{intermediate\_checks\}
\item Existing Assertions: \{assertions\}
\end{itemize}

\textbf{Output Requirements:}
Generate a JSON object containing:
\begin{enumerate}
\item \texttt{success\_criterion}: Concise success determination description
\item \texttt{intermediate\_checks}: List of intermediate state checkpoints (retain and optimize original content)
\item \texttt{assertions}: List of programmable assertions (using pseudo-code function format)
\item \texttt{expected\_final\_state}: Expected final system state
\end{enumerate}

\textbf{Assertion Function Examples:}
\begin{itemize}
\item \texttt{file\_exists("path/to/file.xlsx")} - Check if file exists
\item \texttt{file\_contains("file.xlsx", "keyword")} - Check file content
\item \texttt{window\_active("Application Name")} - Check if window is active
\item \texttt{email\_sent\_to("recipient@email.com")} - Check email sent
\item \texttt{clipboard\_contains("text")} - Check clipboard content
\item \texttt{browser\_url\_matches("pattern")} - Check browser URL
\item \texttt{application\_running("app\_name")} - Check application running status
\item \texttt{error\_dialog\_shown()} - Check error dialog (for L4 tasks)
\end{itemize}

\textbf{For L4 tasks (impossible tasks), assertions should verify that the Agent correctly reported the error.}

Output ONLY the JSON object in the following format:
\{
  "success\_criterion": "...",
  "intermediate\_checks": ["Step 1 completion state", "Step 2 completion state", ...],
  "assertions": ["...", "..."],
  "expected\_final\_state": "..."
\}
}
\caption{Prompt for the Metric Refiner node in the refiner pipeline. This module generates programmable assertions for automated evaluation.}
\label{fig:prompt-metric-refiner}
\end{figure*}

\subsection{Prompt for Environment Generator}

In Figure~\ref{fig:prompt-env-unified}, we present the system prompt used in Environment Generator.

\begin{figure*}[p]
\centering
\promptbox{
\textbf{\# Environment File Content Generator}

You are a professional file content generator responsible for creating environment files that support GUI task execution in WindowsWorld.

\textbf{[Global Requirements]}
\begin{itemize}
    \item Generated content MUST strictly satisfy the task instruction.
    \item If the instruction mentions specific entities (e.g., departments, names, thresholds), the content MUST include them.
    \item Output ONLY file content. Do NOT include explanations or metadata.
\end{itemize}

\hrule
\vspace{0.4em}

\textbf{(1) Text-Based Files (txt, md, json, py)}

\textbf{Purpose:} Generate executable or readable text content (e.g., config files, scripts, notes).

\begin{itemize}
    \item Ensure correct syntax for code or structured formats (e.g., valid JSON).
    \item Content must directly support task execution.
\end{itemize}

\vspace{0.3em}
\textbf{Inputs:} Filename, File Type, Content Requirements, Task Instruction

\hrule
\vspace{0.4em}

\textbf{(2) Spreadsheet Files (Excel / CSV)}

\textbf{Purpose:} Generate tabular data for filtering, aggregation, or analysis tasks.

\begin{itemize}
    \item Output MUST be valid JSON with \texttt{"columns"} and \texttt{"data"} fields.
    \item Data MUST include both matching and non-matching records for specified conditions.
    \item Recommended size: 15--25 rows.
\end{itemize}

\vspace{0.3em}
\textbf{Inputs:} Filename, File Type, Content Requirements, Task Instruction

\hrule
\vspace{0.4em}

\textbf{(3) Word Documents}

\textbf{Purpose:} Generate structured document content for reporting or writing tasks.

\begin{itemize}
    \item Use Markdown-style formatting (headings, lists, emphasis).
    \item Ensure all required content specified in the instruction is included.
\end{itemize}

\vspace{0.3em}
\textbf{Inputs:} Filename, File Type, Content Requirements, Task Instruction
}
\caption{
Unified prompt for environment file generation in WindowsWorld.
A single generator produces text files, spreadsheets, and documents with task-aligned content,
ensuring that environment artifacts are executable, verifiable, and consistent with task instructions.
}
\label{fig:prompt-env-unified}
\end{figure*}

\end{document}